\documentclass{article}

\usepackage{arxiv}
\usepackage[utf8]{inputenc} % allow utf-8 input
\usepackage[T1]{fontenc}    % use 8-bit T1 fonts
\usepackage{hyperref}       % hyperlinks
\usepackage{url}            % simple URL typesetting
\usepackage{booktabs}       % professional-quality tables
\usepackage{amsfonts}       % blackboard math symbols
\usepackage{nicefrac}       % compact symbols for 1/2, etc.
\usepackage{microtype}      % microtypography
\usepackage{lipsum}
\usepackage{amsmath}
\usepackage{bm}
\usepackage{url}
\usepackage{stackengine}
\usepackage{caption}
\usepackage{makecell}
\usepackage{multirow}
\usepackage{mwe}
\usepackage{subcaption}
\usepackage{graphicx}
\usepackage{diagbox}
\usepackage{soul,color}
\usepackage[ruled,vlined]{algorithm2e}
\title{Graph Neural Network Surrogate for Seismic Reliability Analysis of Highway Bridge Systems}

\author{
  Tong Liu\\
  Department of Civil and Environmental Engineering\\
  University of Illinois at Urbana-Champaign\\
  Urbana, IL 61801 \\
  \texttt{tongl5@illinois.edu} \\
   \And
 Hadi Meidani \\
  Department of Civil and Environmental Engineering\\
  University of Illinois at Urbana-Champaign\\
  Urbana, IL 61801 \\
  \texttt{meidani@illinois.edu} \\
}

\begin{document}
\maketitle

\begin{abstract}
Rapid reliability assessment of transportation networks can enhance preparedness, risk mitigation, and response management procedures related to these systems. Network reliability analysis commonly considers network-level performance and does not consider the more detailed node-level responses due to computational cost. In this paper, we propose a rapid seismic reliability assessment approach for bridge networks based on graph neural networks, where node-level connectivities, between points of interest and other nodes, are evaluated under probabilistic seismic scenarios. Via numerical experiments on transportation systems in California, we demonstrate the accuracy, computational efficiency, and robustness of the proposed approach compared to the Monte Carlo approach.
\end{abstract}

% keywords can be removed

\section{Introduction}
Extreme hazards continue to influence the infrastructure system,  with losses exceeding hundreds of billions of dollars. For instance, the overall cost to repair or replace the infrastructure system during Hurricane Katrina in 2005 was over \$1 billion \cite{padgett2008bridge}. Furthermore, the transportation system is one of the 16 critical infrastructure sectors according to Cyber Infrastructure Security Agency \cite{cisa2020critical}. The highway system in the US, for instance, encompasses over 164,000 miles in the National Highway System (NHS). The society also relies on the highway system during emergencies to access critical facilities, including hospitals, airports, fire stations, etc. Highway bridges are critical and yet vulnerable components of the transportation infrastructure. The United States has a national inventory of almost 600,000 highway bridges, many of which are deteriorated and considered structurally deficient \cite{congress2022challenges}. Maintaining a reliable highway bridge system can be properly achieved via a framework that quantitatively assesses highway reliability in an accurate and computationally tractable way. 

In the aftermath of natural hazards, the objectives of a reliability assessment framework are to identify the risk potentials to residents and the infrastructure system \cite{zhou2018structural}; to adjust the emergency plan beforehand to reduce the potential losses; to allocate the resources effectively for recovery. A review of the reliability analysis of transportation systems can be found in \cite{Wan2018Jul}. As an example, \cite{rangra2015study} builds a framework considering the human factor in the transportation reliability analysis and then identifies the need for both transportation systems and driving assistance systems. Among quantitative approaches, various metrics are proposed to the reliability evaluation of the transportation system, including connectivity \cite{lian2021adjustment,chen2021bridge}, resilience index \cite{dong2017probabilistic}, and travel time \cite{chen2022seismic}, traffic flow \cite{liu2023heterogeneous}, transportation equity \cite{liu2023optimizing}. Even though the effectiveness of the simulation-based approaches, the computational time for large infrastructure networks is expensive. To address this issue, surrogate modeling has been proposed for network reliability analysis. For instance, \cite{nabian2018deep} provided a deep learning framework to accelerate seismic reliability analysis of a transportation network where the $k$-terminal connectivity measure was used. As another example, \cite{yoon2020accelerated} proposed a neural network surrogate for system-level seismic risk assessment of bridge transportation networks using total system travel time as the evaluation metric. 

Even though surrogate modeling has been proven to provide an effective way for reliability analysis, it has several limitations. Most of the aforementioned studies only offer a graph-level evaluation of road network resilience performance. Furthermore, these neural network surrogates are unable to generalize to different network topologies. In this paper, we seek to compute a more detailed generalized node-level evaluation measure in an efficient way using a neural network-based analysis framework. Specifically, the model evaluates the connectivity probability for all origin-destination pairs after earthquakes using a graph neural network (GNN) model. The major contributions of this work are as follows: (1) to the best of authors' knowledge, this is the first work that evaluates the connectivity for all origin-destination pairs in the road network using the graph neural network; (2) the proposed end-to-end structure of the model achieves high efficiency by avoiding extensive sampling otherwise used in Monte Carlo simulations (MCS) which reduces the computational time; and (3) the proposed model has the ability of inductive learning, i.e., it can predict node connectivities in unseen graphs. We will numerically show that the proposed framework can effectively accelerate the connectivity reliability analysis of highway bridge networks with a case study in the Bay Area in California.

The remainder of this article is structured as follows. A general simulation-based framework for connectivity analysis of transportation networks is described in Section \ref{sec:bridgeconnectivityanalysis}. Then, Section \ref{sec:gnnconnectivityanalysis} presents the node-level bridge connectivity analysis for earthquake events using a graph neural network. Furthermore, a case study of the highway bridge system in California is presented to demonstrate the accuracy and efficiency of the proposed framework in Section \ref{sec:experimentandresults}. Finally, the discussion and conclusion of the proposed framework are presented in Section \ref{sec:conclusion}.

\section{Transportation System Reliability Analysis}
\label{sec:bridgeconnectivityanalysis}

A transportation network can be represented by a graph $G = (V, E)$ where $V$ and $E$ denote the sets of nodes and edges between these nodes, respectively. In a highway bridge system, a node is a highway intersection, and an edge is a highway segment between two intersections. Each link is denoted by $(u,v) \in E$ with $u$ and $v$ being the indices of its two end nodes. In this work, we assume the bridge is the only component that can fail in the network due to an earthquake, and an edge $(u,v)$ is removed from the graph only because of a bridge failure.  

Given a source node $s$ and a target node $t$, $s$ and $t$ are connected when there exists at least one active path $l_{s,t} = \{({s,v_1}), ({v_1,v_2}), ({v_1,v_2}), \dots, ({v_n,t})\}$ between $s$ and $t$. Let us denote the the set of all possible paths as the set $L_{s,t} = \{l^i_{s,t}\}_{i = 1}^{n_{s,t}}$ where $n_{s,t}$ is the total number of active path between $s$ and $t$. We seek to compute the node-to-node connectivity (as a probability) using the survival probability of the bridges in the network.

To do so, we first model the survival state of the path $i$ between $s$ and $t$ as a binary Bernoulli variable given by
\begin{equation}
\label{eq:linkprob}
x_{s,t,i} = \left\lbrace\begin{array}{cl}1, &\quad \mathrm{with\ probability\ } p_{s,t,i},\\ 0, &\quad \mathrm{with\ probability\ } 1-p_{s,t,i},\end{array}\right.
\end{equation}
where $\{1,0\}$ denotes the survived and failed states, respectively. A path with at least one failed bridge will be removed. If path $i$ between $s$ and $t$ consist of $m_{s,t,i}$ bridges, the survival probability of  the path  can be represented with
\begin{equation}
p_{s,t,i} = \prod_{j=1}^{m_{s,t,i}} p_{s,t,i,j},
\label{eq.pathsurv}
\end{equation}
where $p_{s,t,i,j}$ is the survival probability of the $j$th bridge  on   path $i$ between $s$ and $t$. The path failure probability is calculated by subtracting the survival probability from one. The detail of calculating bridge survival probability is shown in Section \ref{sec:module1}. Because of the numerous path candidates, it is computationally intractable to compute the node-to-node connectivity probability directly given bridge and path failure probabilities. As a result, a Monte Carlo (MC) approach is adopted in this work, where realizations of the network are obtained by randomly removing paths based on path survival probabilities $p_{s,t,i}$ calculated using Equation~\ref{eq.pathsurv}. Specifically, for the $k$th MC realization of the network with the removed paths, we check all the possible paths between a given pair $s$ and $t$ using the breadth-first search (BFS) algorithm with time complexity of $\mathcal{O}(|\mathcal{V}|+|\mathcal{E}|)$. Then, the binary node-to-node connectivity $p_{s,t}^k$ is set equal to one, if there is at least one path between $s$ and $t$, and it is set to zero, otherwise. With $N$ MC samples, the none-to-node connectivity probability can be approximated as
\begin{equation}
P_{s,t} = \frac{1}{N}\sum_{k=1}^Np^k_{s,t}.
\end{equation}
Enough MC samples are used until the quantity of interest converges, which is set to be when the standard deviation of connectivity probability becomes less than 0.01.

\section{Graph Neural Network for Node-Level Transportation System Reliability Analysis}
\label{sec:gnnconnectivityanalysis}
\subsection{Graph Neural Network}
\label{sec:gnn}

In most of the first applications of neural networks, the input data is structured and typically of Euclidean structure. In graph applications, on the other hand, input is non-Euclidean and unstructured, containing different types of topologies. As such, normal neural network frameworks cannot handle graph data since the structure of the input data is not fixed. This paper leverages the GNN to tackle the non-Euclidean graph-structured inputs. Using the previously introduced notation for the graph, $G = (V, E)$, features used in the GNN consist of node features $X_n \subset \mathbb{R}^{|V|\times F_n}$ and edge features $X_e \subset \mathbb{R}^{|E|\times F_e}$, where $|V|$ and $|E|$ denote the number of nodes and edges, $|F_v|$ and $|F_e|$ denote the dimension of features for each node and edge. The process of combining features from a node and other nodes is called message passing. There are various approaches to exploiting node features. For instance, graph convolutional network \cite{Kipf2016Sep} accomplishes message passing by using the adjacency matrix $A$, or in an improved way by using a normalized adjacency matrix. 

However, the adjacency matrix contains the intra-node information, and it can only be used for transductive learning tasks. To overcome this obstacle, GraphSAGE \cite{hamilton2017inductive} is proposed to enable inductive learning tasks without exploiting the adjacency matrix. GraphSAGE uses parametrized neural networks to aggregate node information based on the central vertex and its neighbors. The learned node aggregation consists of two stages: feature aggregation and feature update. {The learned aggregation step is modeled as a single layer in the graph neural network.}

\begin{figure}[hbt]
\centering
\includegraphics[width=\textwidth]{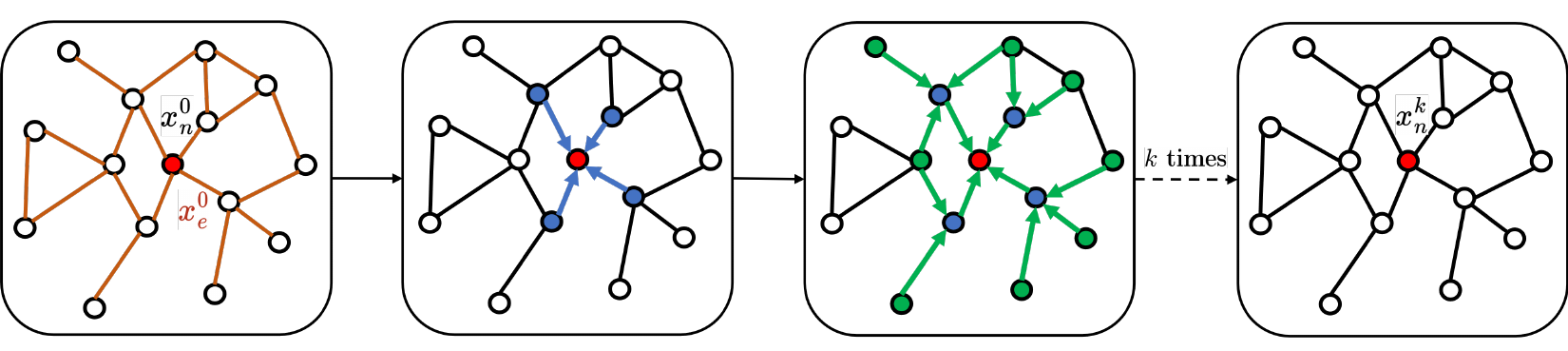}
\caption{Message-passing with both node and edge features.}
%  Left figure represents  original graph at step 0, where node features $x_n^0$ and edge features $x_e^0$ are unrelated. Second figure from left shows the passing of node features  and edge features (shown with arrows) in Step 1 passings (in blue) over 1-hop neighbors. Third figure shows Step 2 passings (in green)  over 2-hop neighbors. Right figure shows the graph after $k$ steps of message passing.
\label{fig:gnn}
\end{figure}

In the feature aggregation stage, {at step $k$ (or in the $k^{\mathrm{th}}$ layer) for each node $v$}, we aggregate the features of its neighbors denoted by  $x_u^k \in \mathbb{R}^{1\times d_n^k}, \forall u \in \mathcal{N}(v)$  according to:
\begin{equation}
\label{eq:node_aggr}
x^{k+1}_{\mathcal{N}(v)} = f\left(\{x_u^k, \forall u \in \mathcal{N}(v)\}\right),\ \forall v \in V,
\end{equation}
where $f$ is the aggregation function, here chosen to be the mean aggregator function, and $d_n^k$ is the node embedding dimension at layer $k$. Then in the update stage at step $k$ for each node $v$, the updated node features $x_v^{k+1}$ will be computed using the previous node features $x_v^k \in \mathbb{R}^{1\times d_n^k}$ and the aggregated features $x^{k+1}_{\mathcal{N}(v)} \in \mathbb{R}^{1\times d_n^{k+1}}$ according to:
\begin{equation}
\label{eq:normal_update}
x^{k+1}_{v} = \sigma \left(g\left(\phi\left(x^k_v, x^{k+1}_{\mathcal{N}(v)}\right); W\right)\right), \ \forall v \in V,
\end{equation}
where $\phi$ is the concatenating function, $\sigma$ is a nonlinear activation function, $g(\cdot;W)$ represents a fully connected layer with parameter $W$. In this work, we extend the learning aggregation process in \cite{hamilton2017inductive} to include both node and edge features. At step $k$, edge features $x_e^k \in \mathbb{R}^{1\times d_e^k}$ of the edge connecting each node $v, \forall u \in \mathcal{E}(v)$, are aggregated through

\begin{equation}
\label{eq:edge_aggr}
x^{k+1}_{\mathcal{E}(v)} = f\left(\{x_e^k, \forall e \in \mathcal{E}(v)\}\right),\ \forall v \in V.
\end{equation}

where $d_e^k$ is the edge embedding dimension at layer $k$. The aggregation step for node embedding is the same as Equation \ref{eq:node_aggr}. Then in the update step, both node features and edge features are passed through a neural network:
\begin{equation}
\label{eq:modify_update}
x^{k+1}_{v} = \sigma \left(g\left(\phi\left(x^k_v, x^{k+1}_{\mathcal{N}(v)}, x^{k+1}_{\mathcal{E}(v)}\right); W\right)\right), \ \forall v \in V.
\end{equation}
By repeatedly using Equations \ref{eq:node_aggr}, \ref{eq:edge_aggr} and \ref{eq:modify_update}, the node and edge features of multiple-hop neighbors of the central node are aggregated into the features of that central node. Figure~\ref{fig:gnn} illustrates the process of message-passing when both node  and edge features are passed. The left figure represents the original graph at step 0 where the node features $x_n^0$ and edge feature $x_e^0$ are initialized. Then in the aggregation step, which is shown in the middle figure, the node feature and edge feature are passed following the arrow direction. Taking the red node as an example, in step 1, the node features and edge features with blue color are passed into the red node and then updated, which are 1-hop neighbors. Then in step 2, the node features and edge features with green color are passed into the red node and then updated, which are 2-hop neighbors. After the k-step update, the node feature $x_n^k$  includes  the information from all k-hop neighbors, which is shown in the right figure. Then the node embedding can be further used for node regression and other downstream tasks.

\subsection{Graph Neural Network for Bridge Connectivity Analysis}
\label{sec:gnn_bridge}
\begin{figure}[hbt]
\centering
\includegraphics[width=\textwidth]{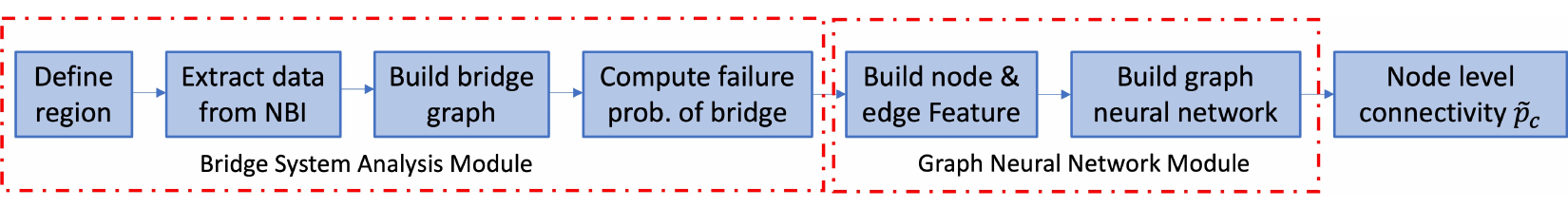}
\caption{Pipeline of connectivity analysis of highway bridge system.}
% It consists of two modules: bridge system analysis module and graph neural network module.
\label{fig:pipeline}
\end{figure}

The pipeline of graph neural network surrogate for bridge connectivity analysis is shown in Figure \ref{fig:pipeline}. The pipeline consists of two major components: a bridge seismic analysis module and a graph neural network module.

\subsubsection{Bridge Seismic Analysis Module}
\label{sec:module1}
This module consists of bridge graph generation and bridge failure probability calculation. To generate the graph, we first need to define a region of interest for reliability analysis. The edges in the graph are highways, freeways, and expressways within the region of interest. The intersections of these roadways are considered as nodes in the graph. To calculate the bridge failure probability, the first step is to extract bridge information in the considered highway system, which could be obtained from the National Bridge Inventory (NBI) \cite{NBI}. The useful information for bridge failure probability calculation includes built year, structural material, structural type, number of spans, maximum span length, bridge length, and skew angle. It should be noted that in general, an edge may consist of zero, one, or more than one bridge. 

The second step for bridge failure probability calculation is to determine the ground motion at each bridge site. In this paper, the ground motion prediction equation (GMPE) is adopted, which predicts the characteristics of ground motion, including peak ground acceleration (PGA), spectral acceleration (SA), and its associated uncertainty \cite{stewart2015selection,bommer2010selection}. In the past decades, hundreds of GMPEs have been proposed for predicting PGA and SA. In this work, Graizer-Kalkan GMPE (GK15) \cite{Graizer2016Apr} is adopted. The updated ground-motion prediction model for PGA has six independent predictor parameters: moment magnitude $M$, the closest distance to fault rupture plane in kilometers $R$, average shear-wave velocity in the upper 30m $V_{\mathrm{S30}}$, style of faulting $F$, regional quality factor $Q_0$, and basin depth under the site $B_{\mathrm{depth}}$. In GK15, the peak ground acceleration is calculated as a multiplication of a series of functions. In the natural logarithmic scale, it is given by:

\begin{equation}
\label{eq:gk15_pga}
\ln(\mathrm{PGA}) = \ln(G_1) + \ln(G_2) + \ln(G_3) + \ln(G_4) + \ln(G_5) + \sigma_{\ln(\mathrm{PGA})}, 
\end{equation}
where $G_1$ is a scaling function for magnitude and style of faulting, $G_2$ models the ground-motion distance attenuation, $G_3$ adjusts the distance attenuation rate considering regional anelastic attenuation, $G_4$ models the site amplification owing to shallow site conditions,  $G_5$ is a basin scaling function, and $\sigma_{\ln(\mathrm{PGA})}$ represents variability in the ground motion.

Given the variability in seismic resistance among the bridges in the network, there is insufficient data to conduct a thorough seismic analysis for each individual bridge in the transportation network.  Instead, to assess our proposed model which is focused on the network-level response, without loss of generality,  the failure probability of a bridge is computed using  fragility functions following the guideline of HAZUS-HM \cite{federal2022hazus}. HAZUS-HM categorizes the bridges into 28 primary types based on materials, structure type, build year, number of spans, etc. To estimate the probability of failure of bridges, the geographical location of the bridge, SA at 0.3 s and 1.0 s at the bridge location, and soil condition are also required. The form for 5\% damped spectral acceleration at spectral period T is:

\begin{equation}
\label{eq:gk15_sa}
S_{a,T} = \mathrm{PGA}\times\mu\left(T, M, R, V_{\mathrm{S30}}, B_{\mathrm{depth}}\right), 
\end{equation}
where $\mu$ is the spectral shape function. In HAZUS-HM, the failure probabilities of bridges are computed using the fragility curve. Five damage states are defined: none, slight, moderate, extensive, and complete damage states. For bridges, extensive damage in HAZUS-HM is defined by shear failure, major settlement approach, or the vertical offset of the abutment. In this work, we assume the bridges will stop functioning and be removed from the network when bridge damage is beyond the extensive damage state. The fragility curve for the bridge component is modeled as log-normal distribution functions characterized by median and dispersion. Furthermore, the failure probability of the $j$th edge on path $i$ between $s$ and $t$,  $p_{s,t,i,j} \in [0,1]$  is computed based on the assumption that the edge fails when at least one bridge fails.

\subsubsection{Bridge Connectivity Analysis Module}
\label{sec:module2}
To build the training data and testing data for graph neural networks, we need first to create node feature and edge feature using the failure probability computed in \ref{sec:module1}. The edge feature is the failure probability of each edge, which has already been computed. It should be noted that in conventional connectivity analysis, there is no feature directly assigned to the nodes, i.e., the intersection of the roadways. In this work, however, we consider node features in the graph neural network models. Four local and global graph characteristics are considered as features of node $v$: the degree of the node $\mathrm{deg}(v)$; the largest failure probability of edges connected to the node $\max(\{p(e), \forall e \in \mathcal{E}(v)\})$; the smallest failure probability of edges connected to the node $\min(\{p(e), \forall e \in \mathcal{E}(v)\})$ and number of hops on the shortest path to the target node $t$ h(v, t). All the node features in the graph are denoted by $X_n\in \mathbb{R}^{|N|\times F_n}$ with the number of node features $F_n=4$.

\begin{figure}[htb]
\centering
\includegraphics[width=0.7\textwidth]{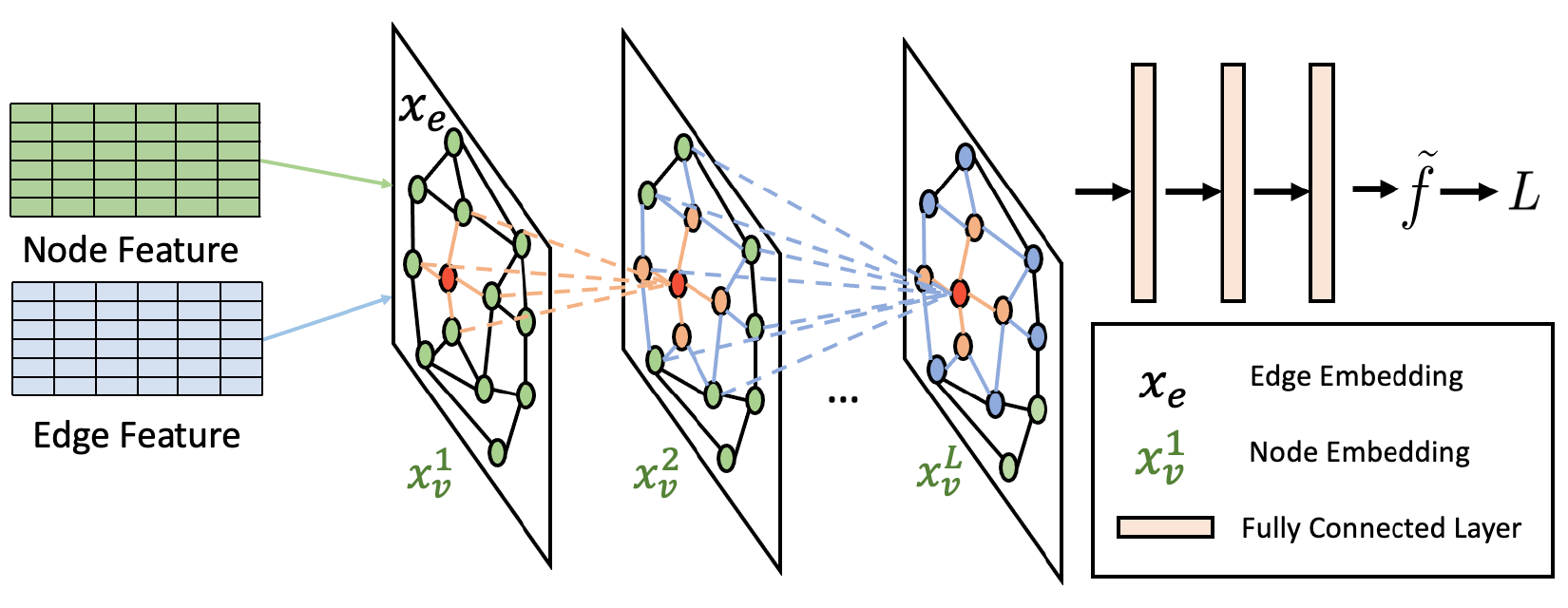}
\caption{Architecture of graph neural network model for node-level connectivity analysis.}
% The model consists of message passing layers, dropout layers, activation layers and fully connected layers. The inputs  are node  and edge features. The model output  is the node-level connectivity.
\label{fig:model}
\end{figure}

The proposed node features characterize each node from several perspectives: the degree represents the possible paths connecting the node to its neighbors; the range of failure probabilities of edges connected to the node represents the distribution of the failure probability; the number of hops represents a global feature characterizing the distance with respect to the  target node. It should be noted that in addition to the range of edge failure probabilities (the largest and smallest failure probabilities of connected edges), one can also consider multiple quantiles to more precisely capture the distribution of failure probabilities for a potentially more accurate analysis.

%Another difficulty in the graph neural network is the over-smoothing issue. In particular, because of the aggregation and update operations in the message passing, the interacting nodes have similar representations and the feature embeddings of different nodes in the deep graph neural network would tend to be similar. To alleviate this issue, differentiable group normalization \cite{zhou2020towards} and dropEdge\cite{rong2019dropedge} were proposed to add normalization into graph neural networks in different forms. In this work, we add a dropout layer after the message-passing layer since  dropout can randomly remove  neurons in the forward propagation. It is equivalent to randomly removing edge features and node features from the original graph in the training process, which can be considered as a normalization strategy.

The architecture of the proposed graph neural network model for bridge connectivity analysis is shown in Figure \ref{fig:model}. A single block of message passing consists of a message-passing layer, a dropout layer, and an activation layer. The skipping layer connection is used in our proposed architecture to mitigate vanishing or exploding gradient problems. A regression block with multiple feed-forward fully connected layers is concatenated after the last message passing block. The output of the model is the node-level connectivity probability $P_c\in \mathbb{R}^{|V|\times1}$. The L1 loss is used for quantifying the difference between the MCS and prediction.

\section{Experiments and Results}
\label{sec:experimentandresults}
\subsection{Experiment Setup}
\label{sec:experimentsetup}
\begin{figure}[hbt]
\centering
\includegraphics[width=0.3\textwidth]{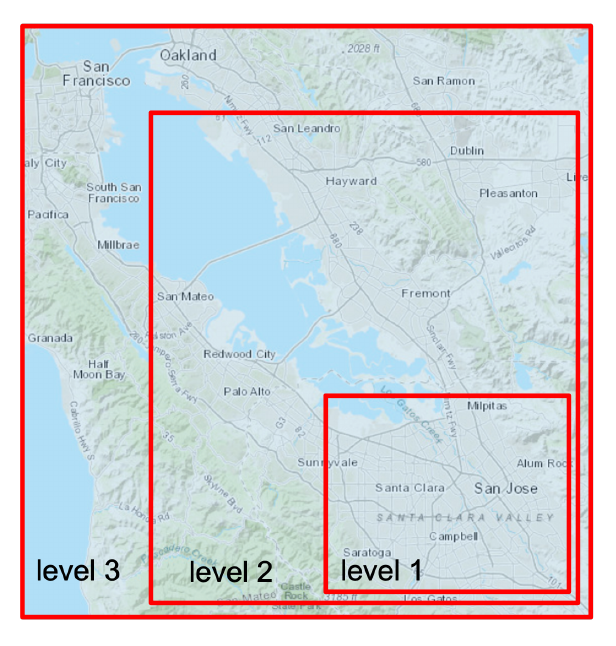}
\caption{Map illustration of the studied regions in different levels.}
\label{fig:levels}
\end{figure}

The proposed GNN-based reliability analysis is applied to the highway bridge system of the California Bay Area. The transportation system contains highways, freeways, and expressways connecting major airports and hospitals. The considered study area includes the major cities of Santa Clara, Mountain View, and San Jose, with a large population base. Therefore, it is important to maintain the connectivity of the highway bridge system after extreme events. In this experiment, three different regions of study are considered (levels 1 to 3). The map illustrations are shown in Figure \ref{fig:levels} obtained from Google Map \cite{googlemap}. The 1989 Loma Prieta earthquake is chosen as the seismic event. The earthquake of scenario $i$ is scaled to different magnitudes:
\begin{equation}
    M_i = M_u - \lambda_i,
\end{equation}
Where $M_u$ is the upper bound of the possible earthquake magnitude and set to be 8.0 in this study, and $\lambda_i$ is a random sample following truncated exponential distribution with the shape parameter of 1.5. Furthermore, in this paper, without loss of generality in evaluating our proposed GNN approach, we assume the seismic intensities at the location of components are perfectly correlated \cite{adachi2009serviceability}, where the residual of PGA will be the same for all the components in the transportation system.  Figure \ref{fig:pipeline} shows the pipeline of data generation. The details of these three regions, including the number of nodes, $N_n$, number of edges, $N_b$, and number of bridges, $N_b$,  are shown in Table \ref{tab:levels}. In the Monte Carlo simulation (MCS) approach, the number of samples $N_s$ is set to 10,000. For each node in the graph, 200 earthquake realizations are generated, where the first 100 realizations are allocated for training and the subsequent 100 realizations are designated for testing. Also, we held out 20\% of the nodes for testing. To ensure that the removed nodes are more or less evenly scattered on the network, the serial graph partitioning algorithm \cite{METIS} is applied for the partitioning of the graph. 

The graph neural network used in this work consists of five graph message-passing layers and three fully connected layers for regression purposes. The dimension of the hidden message passing layer and fully connected layer is 512. The rectified linear unit (ReLU) is chosen as the activation function. The dropout rate is set to 0.1, and the Adam optimizer is used to minimize the L1 loss, which is the mean absolute error, with a learning rate of 0.001. The model is trained using mini-batch training with 200 epochs and a batch size of 64, which is commonly chosen in the neural network training \cite{liu2023physics}. 

\begin{table}[hbt]
\caption{Statistic of different region levels}
\label{tab:levels}
\centering
\small
\renewcommand{\arraystretch}{1.25}
\begin{tabular}{cccc}
    \hline\hline
    Graph Level & \# Node $N_n$ & \# Edge $N_e$ & \# Bridge $N_b$\\
    \hline
    Level 1     & 39        & 64      & 245     \\
    Level 2     & 84        & 133     & 448     \\
    Level 3     & 103       & 159     & 628     \\
    \hline\hline
    \end{tabular}
\normalsize
\end{table}

\subsection{Prediction Results}
\label{sec:predictionresult}
\begin{figure}[hbt]
\centering
\begin{subfigure}{0.47\linewidth}
  \centering
  \includegraphics[width=\linewidth]{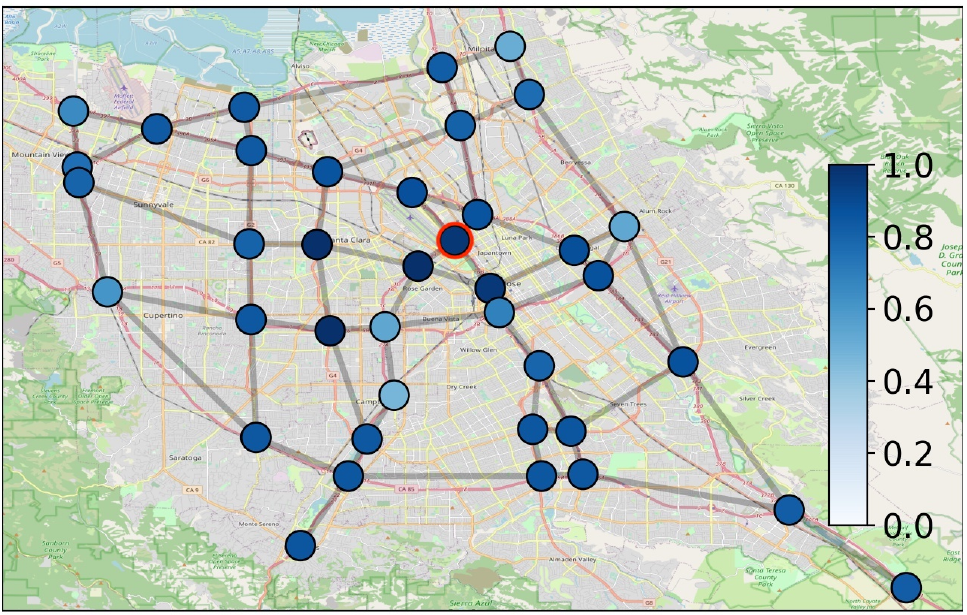}
  \caption{Predicted connectivity probability from GNN}
  \label{fig:level1_reg_pred}
\end{subfigure}
\begin{subfigure}{0.47\linewidth}
  \centering
  \includegraphics[width=\linewidth]{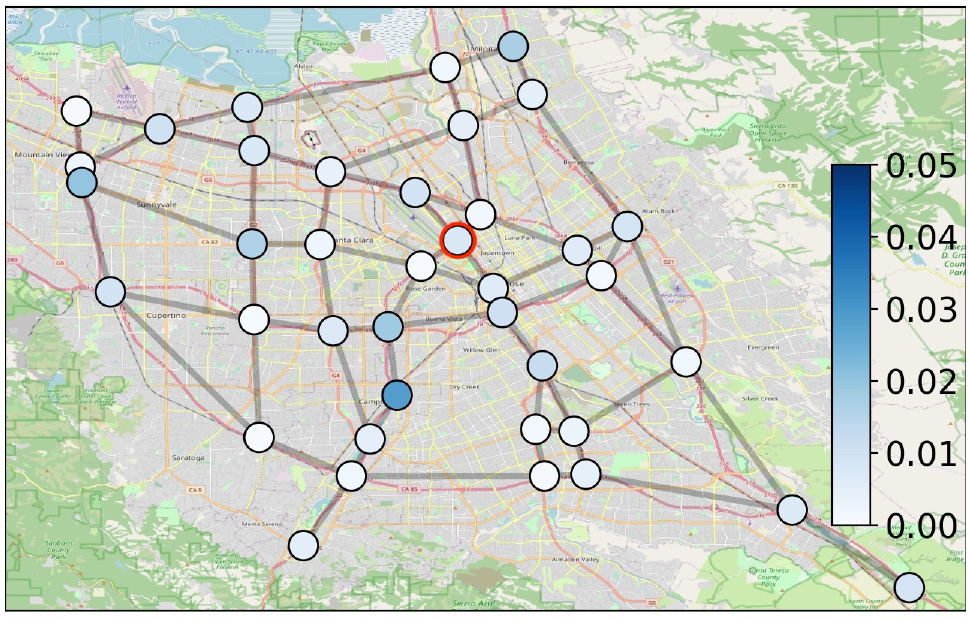}
  \caption{Absolute error  between GNN and MCS predictions}
  \label{fig:level1_reg_diff}
\end{subfigure}
\caption{Prediction of connectivity probabilities between the target node (red circle) and the other nodes in the Level 1 region. The nodes are color-coded based on the predicted connectivity probability (Figure \ref{fig:level1_reg_pred}) and the absolute error between predicted connectivity probabilities from GNN and MCS (Figure \ref{fig:level1_reg_diff}). }
\label{fig:level1_reg}
% . Left figure is the node-level connectivity probability from proposed model; Right figure is the difference between MCS and GNN prediction. The edge label denotes the failure probability of the edge. The darker node color indicates higher connectivity probability between the each node to target node.
\end{figure}

\begin{figure}[hbt]
\centering
\begin{subfigure}{0.47\linewidth}
  \centering
  \includegraphics[width=\linewidth]{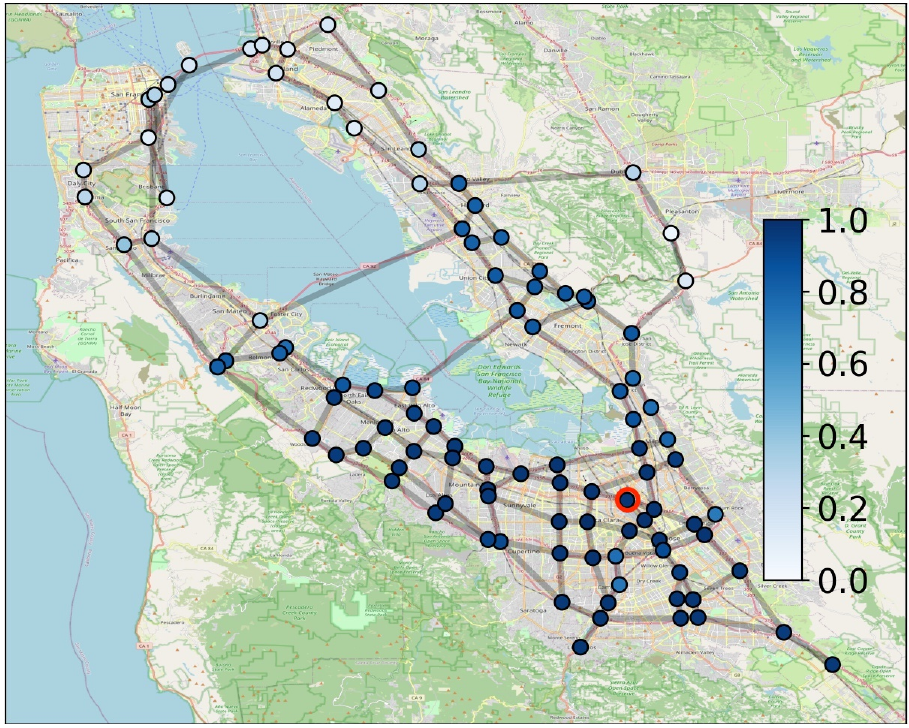}
  \caption{Predicted connectivity probability from GNN}
  \label{fig:level3_reg_pred}
\end{subfigure}
\begin{subfigure}{0.47\linewidth}
  \centering
  \includegraphics[width=\linewidth]{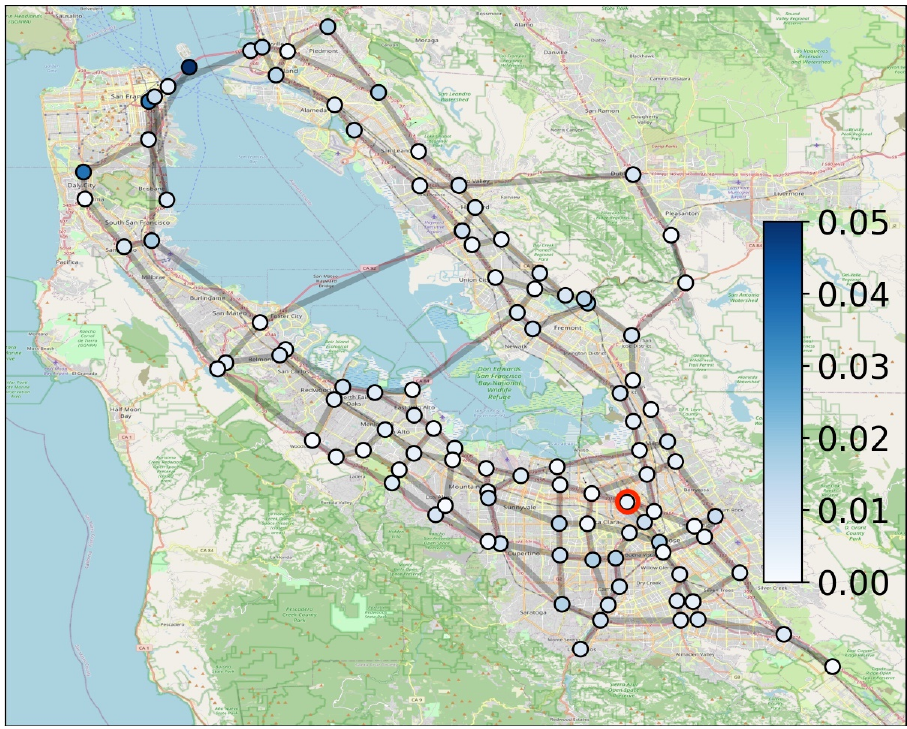}
  \caption{Absolute error between MCS and GNN predictions}
  \label{fig:level3_reg_diff}
\end{subfigure}
\caption{Prediction of connectivity probabilities between the target node (red circle) and the other nodes in the Level 3 region. The nodes are color-coded based on the predicted connectivity probability (Figure \ref{fig:level3_reg_pred}) and the absolute error  between predicted connectivity probabilities from GNN and MCS (Figure \ref{fig:level3_reg_diff}). }
% Left figure is the connectivity probability prediction from graph neural network. Right figure is the MAE between MCS and prediction.
\label{fig:level3_reg}
\end{figure}

We test the performance of the proposed approach from multiple perspectives. First, we consider a regression problem where we calculate the probability of connectivity between a target node and the rest of the nodes in the region. The target node is located in downtown San Jose, which is a critical location for the large population base. We calculate the connectivity probability for three regions of interest. The prediction results for the Level 1 region are shown in Figure \ref{fig:level1_reg}. The maximum and average mean absolute error (MAE) between MCS and GNN prediction is 0.037 and 0.020, respectively. Figure \ref{fig:level3_reg} shows the prediction results and the corresponding error for the Level 3 graph. For this case, the maximum MAE in the graph is 0.051 and the mean prediction MAE is 0.015. This demonstrates the accuracy of the proposed GNN surrogate model.

\begin{figure}[hbt]
\centering
\begin{subfigure}{0.47\linewidth}
  \centering
  \includegraphics[width=\linewidth]{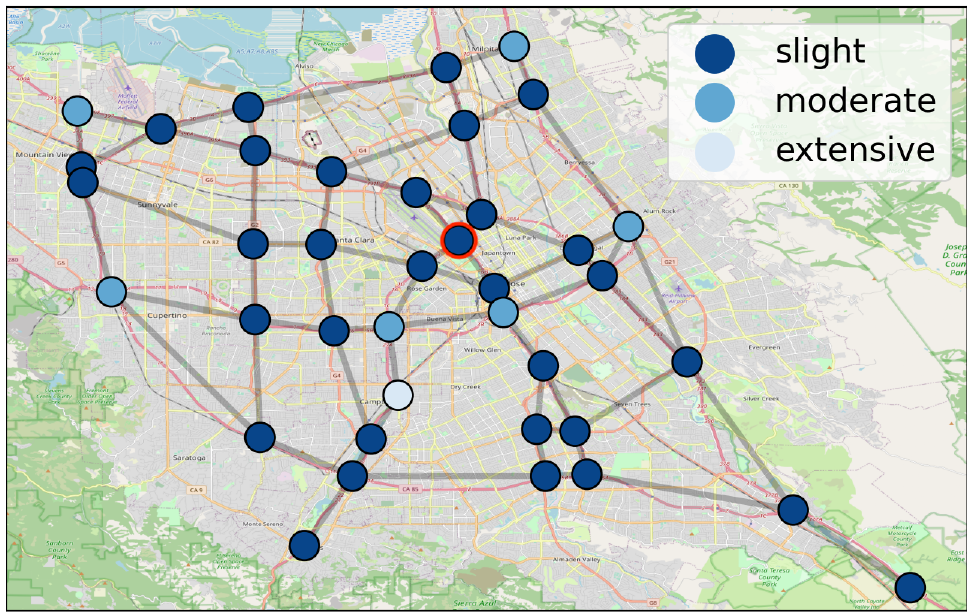}
  \caption{Predicted connectivity classes from MCS}
  \label{fig:level1_cla_gt}
\end{subfigure}
\begin{subfigure}{0.47\linewidth}
  \centering
  \includegraphics[width=\linewidth]{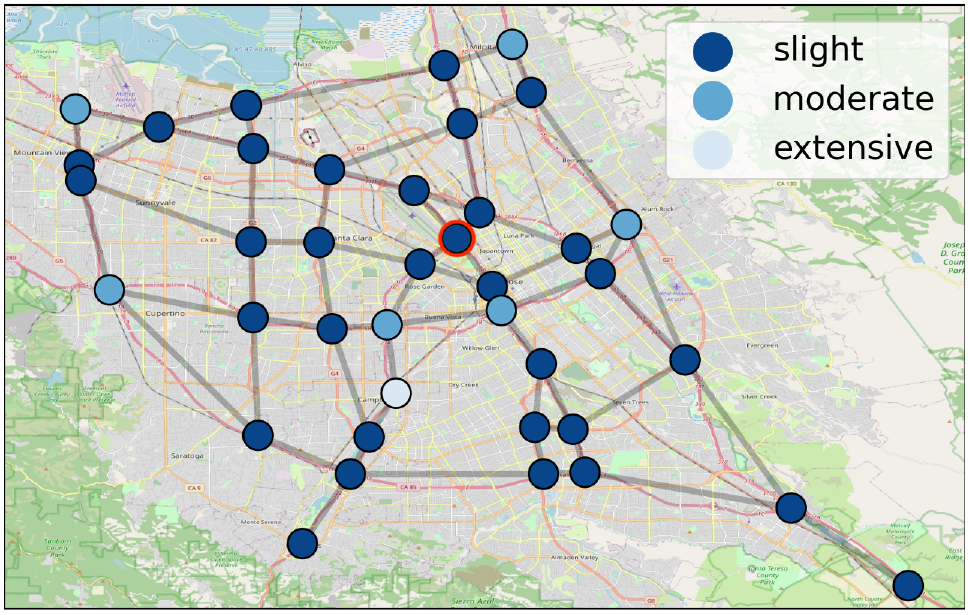}
  \caption{Predicted connectivity class from GNN}
  \label{fig:level1_cla_pred}
\end{subfigure}
\caption{Identical predictions of connectivity classes obtained from MCS (Figure \ref{fig:level1_cla_gt}) and GNN models (Figure \ref{fig:level1_cla_pred}) for the Level 1 region. The red circle shows the target node. Three classes represent normal connection, minor disconnection, and major disconnection.}
\label{fig:level1_cla}
% . The cutoff thresholds are 0.5 and 0.75. Left figure is the connectivity classification from MCS. Right figure is the classification prediction of each node. Yellow, green, purple represent normal operation, minor disconnection and major disconnection, respectively. The edge label denotes the failure probability of the edge.
\end{figure}

Furthermore, from the perspective of decision-making, stakeholders may choose to consider a set of  node connectivity classes. In this case, we evaluate how accurate the predicted classes will be compared to the MCS approach. It should be noted that a different classification model is not trained in this case. We only predict the connectivity classes by assigning the predicted connectivity probability of each node into a class. In this example, three classes for node connectivity probabilities, namely normal connection, minor disconnection and major disconnection, for the connectivity probabilities falling in the ranges [0.75, 1.0], [0.5, 0.75] and [0, 0.5], respectively. Figure \ref{fig:level1_cla} shows the classification result using the same example in Figure \ref{fig:level1_reg}. F1 score is chosen as the classification evaluation metric, which is 1.0 in this example.

Moreover, we conducted an ablation study on hyperparameters to assess their influence in both regression and classification tasks comprehensively. The study encompasses various hyperparameters, including the number of GNN layers, the dimension of the hidden layer, the learning rate during training, and the features employed in the training process. Three metrics including MAE and F1 score of two-class and three-class classification are considered in the ablation study, which is aligned with the aforementioned experiment. Each scenario was subjected to three runs utilizing different random seeds, and the reported results represent the average values over these runs. Table \ref{tab:metric} provides a summarized overview of performance metrics for different regions. Notably, a reduction in the number of GNN layers is associated with a decrease in prediction performance. Furthermore, an increase in the learning rate and a decrease in the dimension of the hidden layer results in less stable training performance, thereby diminishing overall performance. When utilizing solely the edge feature during both training and prediction, it becomes evident that message passing is less effective when compared with the performance of the complete model. This observation underscores the importance of the pipeline's efficiency and the hyperparameter selection. 

The proposed model is also compared with other neural network models including support vector regression (SVR), fully connected neural network (FCNN), graph convolutional network (GCN), and graph attention network (GAT) \cite{velivckovic2017graph}, where the evaluation metrics are the same as the ablation study. The performance comparison over different regions is shown in Table \ref{tab:model}. The fully connected network has a relatively low accuracy due to the reason that it lacks generalization capability to  different graphs, and it cannot learn the graph feature effectively. Compared to GAT and GCN prediction, the proposed model still has a relatively high accuracy in terms of MAE and F1 scores in all three  regions.

\begin{figure}[hbt]
\centering
\begin{subfigure}{0.33\linewidth}
  \includegraphics[width=\linewidth]{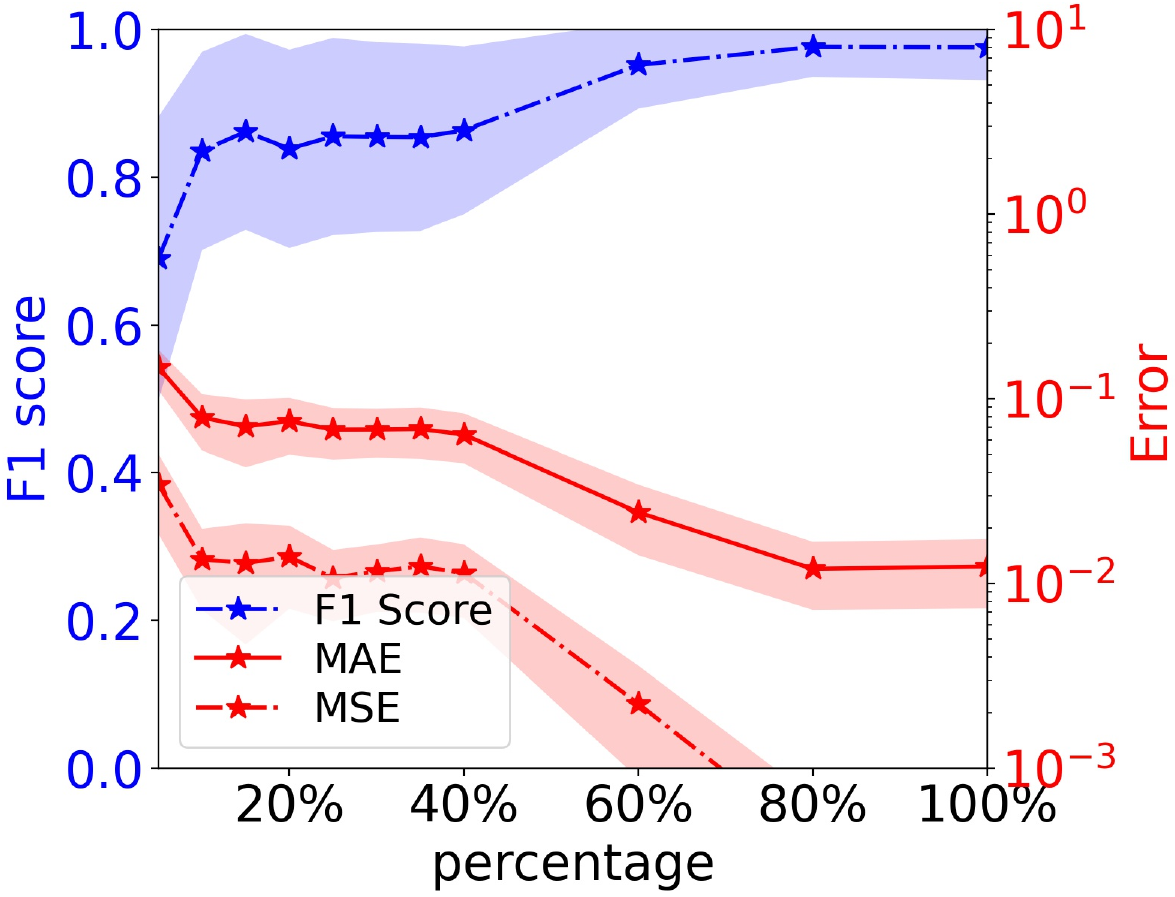}
  \caption{Level 1}
  \label{fig:level1_metric}
\end{subfigure}
\begin{subfigure}{0.33\linewidth}
  \centering
  \includegraphics[width=\linewidth]{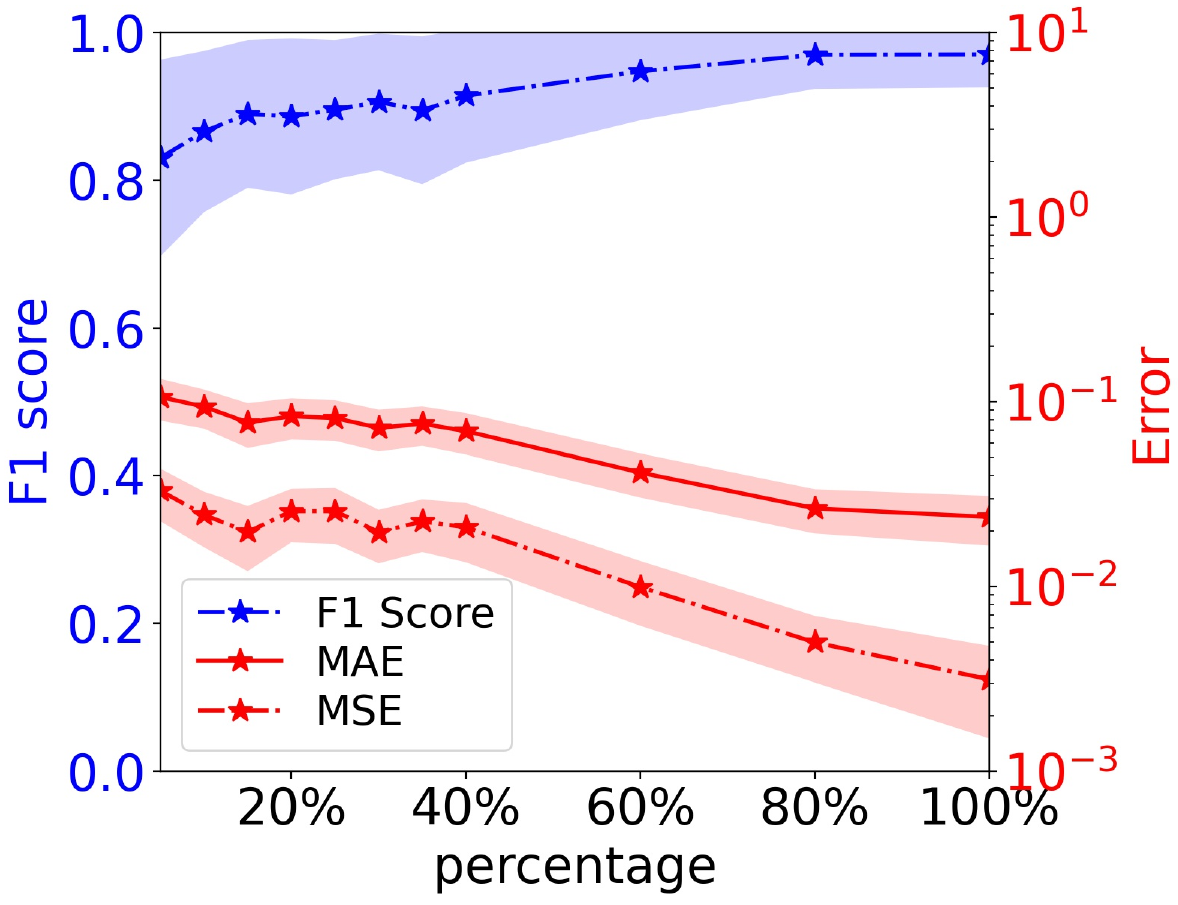}
  \caption{Level 2}
  \label{fig:level2_metric}
\end{subfigure}
\begin{subfigure}{0.33\linewidth}
  \centering
  \includegraphics[width=\linewidth]{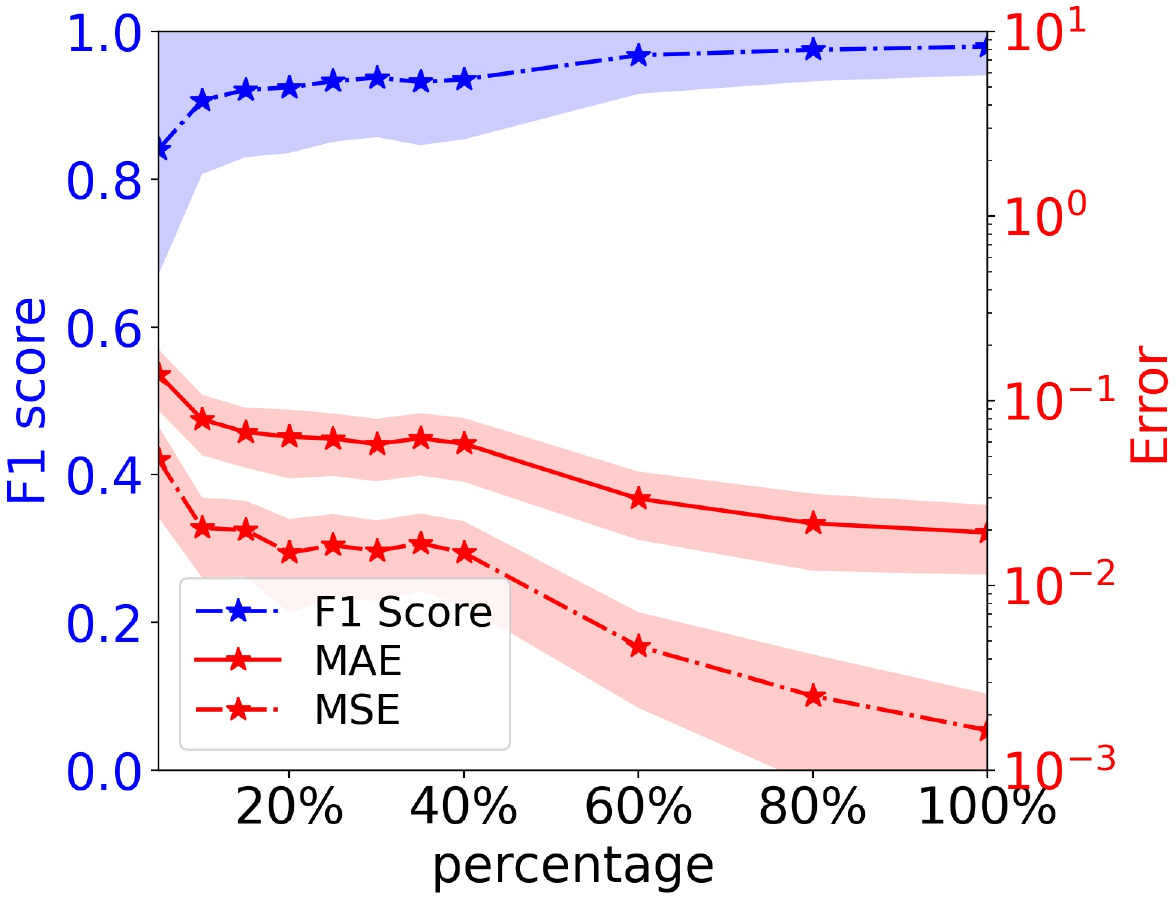}
  \caption{Level 3}
  \label{fig:level3_metric}
\end{subfigure}
\caption{Relationship between different metrics and the percentage of training nodes shown for the Level 1 region (Figure \ref{fig:level1_metric}), Level 2 region (Figure \ref{fig:level2_metric}), and Level 3 region (Figure \ref{fig:level3_metric}).}
% Three metrics are plotted: F1 score, mean absolute error and mean square error. Right $y$-axis is in logarithmic scale.
\label{fig:metric}
\end{figure}

\begin{figure}[hbt]
\centering
\begin{subfigure}{0.33\linewidth}
  \includegraphics[width=\linewidth]{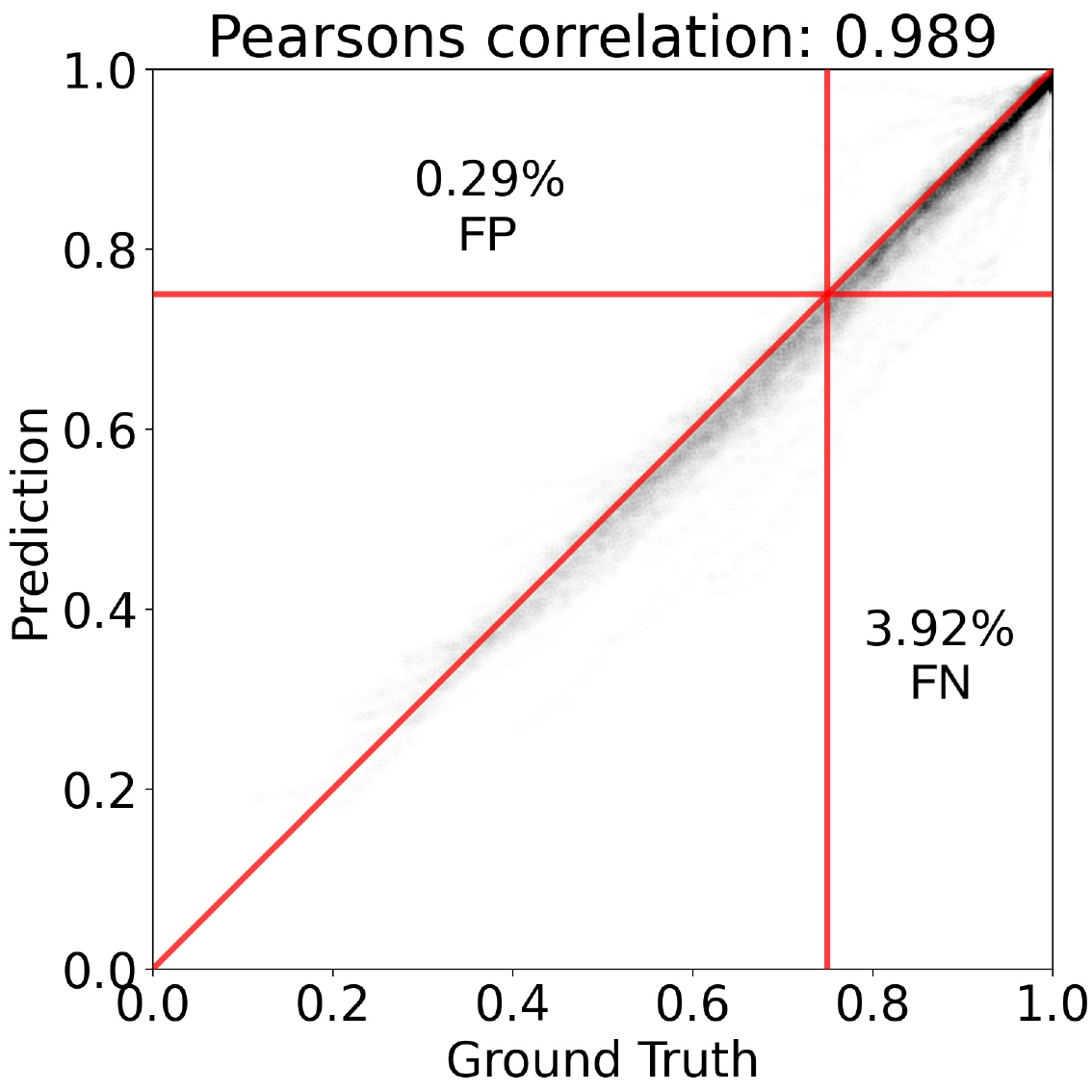}
  \caption{Level 1}
  \label{fig:level1_corr}
\end{subfigure}
\begin{subfigure}{0.33\linewidth}
  \centering
  \includegraphics[width=\linewidth]{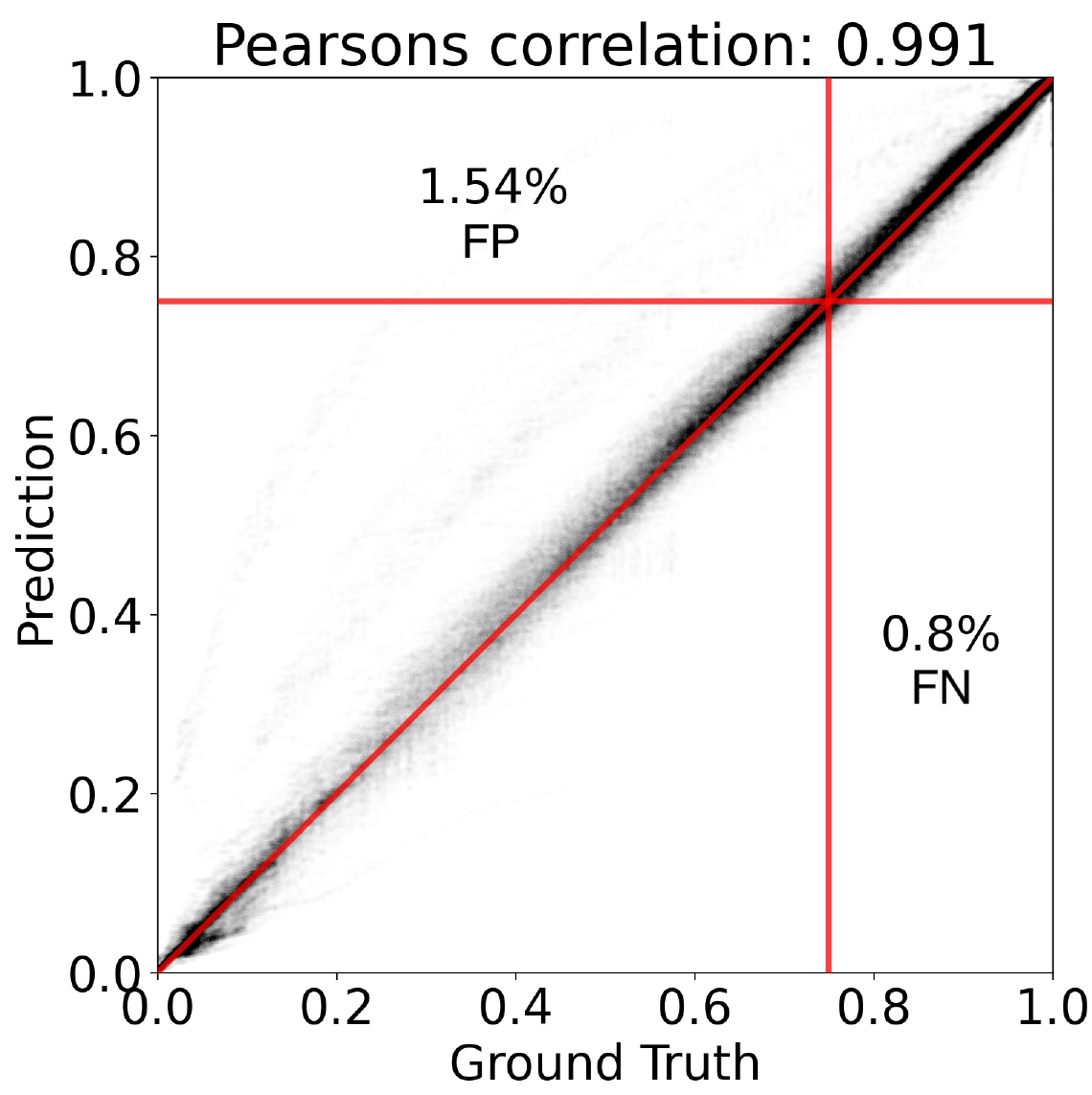}
  \caption{Level 2}
  \label{fig:level2_corr}
\end{subfigure}
\begin{subfigure}{0.33\linewidth}
  \centering
  \includegraphics[width=\linewidth]{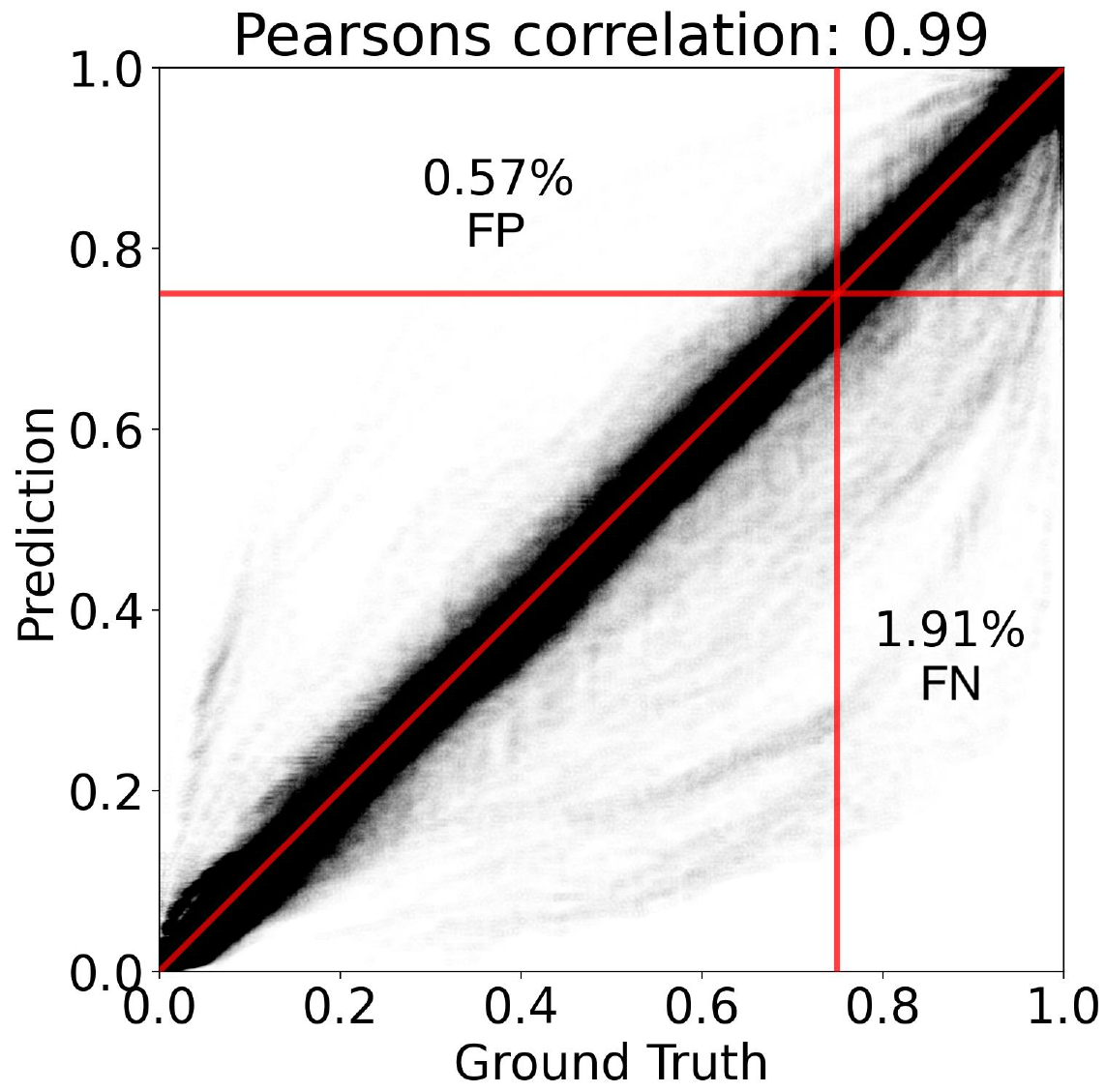}
  \caption{Level 3}
  \label{fig:level3_corr}
\end{subfigure}
\caption{Relationship between predicted node connectivity and MCS predictions shown for the Level 1 region (Figure \ref{fig:level1_corr}), Level 2 region (Figure \ref{fig:level2_corr}), and Level 3 region (Figure \ref{fig:level3_corr}). The percentage in the figure represents the proportion of false positive (FP) and false negative (FN) results.}
% Diagonal red line represents the line $y=x$. 
\label{fig:corr}
\end{figure}

In order to assess the robustness of the proposed approach,  we run several experiments in which the number of training nodes within the graph varies. Specifically, a subset of the nodes is considered as the target nodes in training. For each case, a specific ratio of the graph nodes is selected as the "training" target nodes. These ratios are chosen between 5\% and 80\%.  The evaluation metrics include mean square error, mean absolute error, and F1 score. The performances with different levels are demonstrated in Figure \ref{fig:metric}, which shows that when the ratio of training target nodes is from 20\% to 40\%, the F1 score can reach 0.85 and MAE is less than 0.08. Furthermore, when the training target node ratio is 60\%, the F1 score can reach above 0.9, and MAE is less than 0.01. To further evaluate the robustness of the proposed approach, Figure \ref{fig:corr} compares the MCS and GNN predictions of node connectivity for all earthquake realizations. The percentage in each figure indicates the ratio of false positive (FP) and false negative (FN) samples among all samples with the cutoff threshold 0.75, which is less than 5\%. It can also be seen that the Pearson Correlation coefficients between MCS and GNN prediction for all region levels are higher than 0.93.

% \usepackage{graphicx}
\begin{table}[hbt]
\centering
\caption{Ablation study of hyperparameters on three different regions. Three metrics is considered in the comparison: MAE, F1 score of two-class and three-class classification. The cutoff thresholds for two-class and three-class classification are \{0.75\} and \{0.5, 0.75\}, respectively.}
\label{tab:metric}
\resizebox{\textwidth}{!}{%
\begin{tabular}{cccccccccc}
\toprule[1.5pt]
                     & \multicolumn{3}{c}{Level 1} & \multicolumn{3}{c}{Level 2} & \multicolumn{3}{c}{Level 3} \\ \cmidrule[1pt]{2-10}
Model      & MAE            & F1 (2 class)   & F1 (3 class)   & MAE            & F1 (2 class)   & F1 (3 class)   & MAE            & F1 (2 class)   & F1 (3 class)   \\ \midrule[1.5pt]
\# GNN layer = 3     & 0.047   & 0.905   & 0.876   & 0.049   & 0.925   & 0.870   & 0.051   & 0.939   & 0.889   \\
\# GNN layer = 1     & 0.061   & 0.878   & 0.853   & 0.092   & 0.851   & 0.760   & 0.112   & 0.863   & 0.760   \\
Hidden size = 32     & 0.044   & 0.903   & 0.878   & 0.052   & 0.923   & 0.868   & 0.064   & 0.926   & 0.862   \\
Learning rate = 0.01 & 0.070   & 0.858   & 0.829   & 0.138   & 0.756   & 0.636   & 0.175   & 0.805   & 0.618   \\
No node feature      & 0.065   & 0.869   & 0.838   & 0.094   & 0.844   & 0.760   & 0.138   & 0.831   & 0.715   \\
Full model & \textbf{0.032} & \textbf{0.937} & \textbf{0.921} & \textbf{0.029} & \textbf{0.953} & \textbf{0.919} & \textbf{0.038} & \textbf{0.956} & \textbf{0.919} \\ \toprule[1.5pt]
\end{tabular}%
}
\end{table}

% \usepackage{graphicx}
\begin{table}[hbt]
\centering
\caption{Performance comparison among different models on three regions.  FCNN, GAT and GCN are considered to compare with the proposed model. Three metrics are considered in the comparison: MAE, F1 score of two-class and three-class classification. The cutoff thresholds for two-class and three-class classification are \{0.75\} and \{0.5, 0.75\}, respectively.}
\label{tab:model}
\resizebox{\textwidth}{!}{%
\begin{tabular}{cccccccccc}
\toprule[1.5pt]
     & \multicolumn{3}{c}{Level 1} & \multicolumn{3}{c}{Level 2} & \multicolumn{3}{c}{Level 3} \\ \cmidrule[1pt]{2-10}
Model     & MAE            & F1 (2 class)   & F1 (3 class)   & MAE            & F1 (2 class)   & F1 (3 class)   & MAE            & F1 (2 class)   & F1 (3 class)   \\ \midrule[1.5pt]
SVR & 0.061   & 0.874   & 0.834   & 0.092   & 0.821   & 0.738   & 0.172   & 0.793   & 0.672   \\
FCNN & 0.081   & 0.848   & 0.827   & 0.148   & 0.717   & 0.604   & 0.198   & 0.759   & 0.598   \\
GAT  & 0.079   & 0.834   & 0.775   & 0.079   & 0.869   & 0.782   & 0.083   & 0.901   & 0.816   \\
GCN  & 0.056   & 0.886   & 0.853   & 0.037   & 0.944   & 0.906   & 0.055   & 0.937   & 0.878   \\
Our model & \textbf{0.032} & \textbf{0.973} & \textbf{0.921} & \textbf{0.029} & \textbf{0.953} & \textbf{0.919} & \textbf{0.038} & \textbf{0.956} & \textbf{0.919} \\ \toprule[1.5pt]
\end{tabular}%
}
\end{table}

\subsection{Model Performance Under Special Cases}
\label{sec:special}
\begin{figure}[hbt]
\centering
\begin{subfigure}{0.33\linewidth}
  \includegraphics[width=\linewidth]{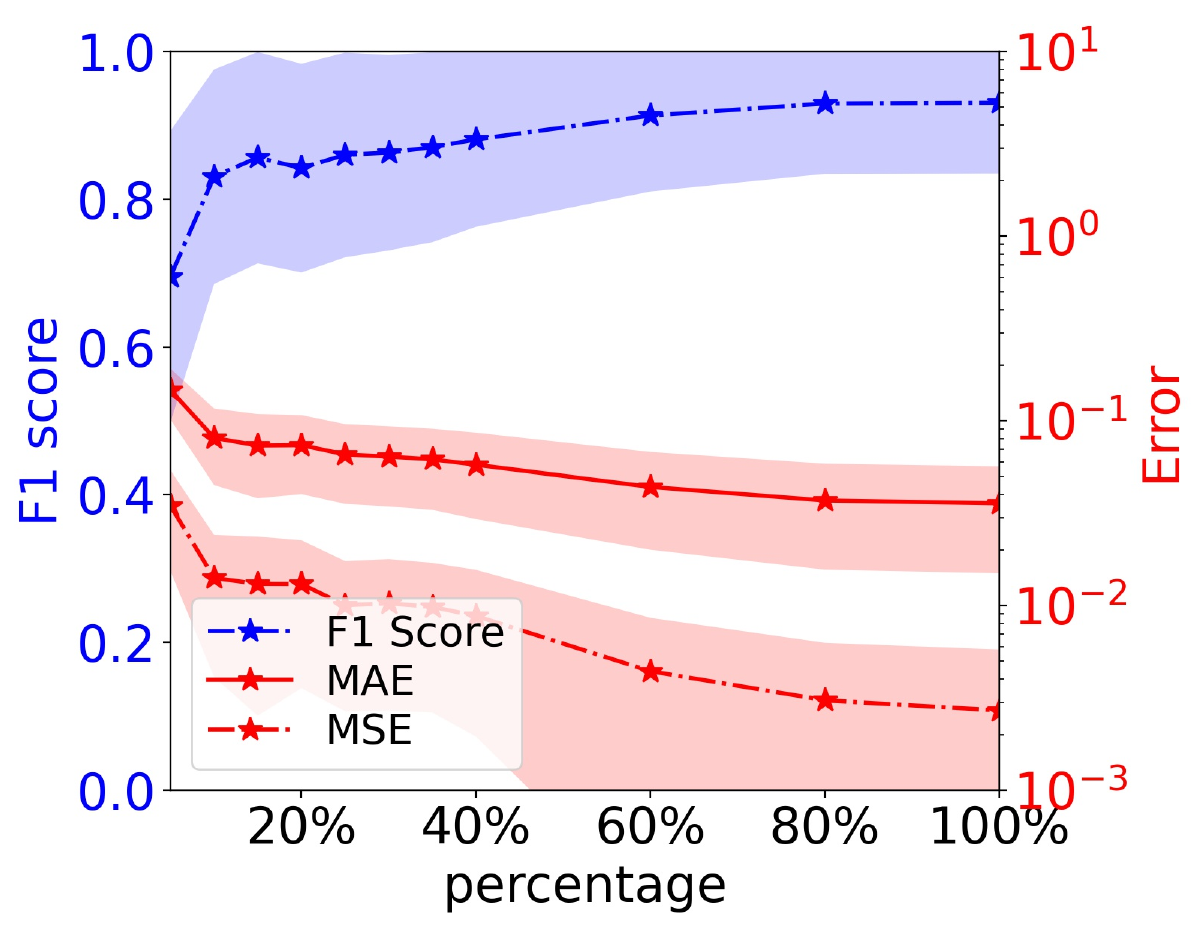}
  \caption{Level 1}
  \label{fig:level1_perturb_metric}
\end{subfigure}
\begin{subfigure}{0.33\linewidth}
  \centering
  \includegraphics[width=\linewidth]{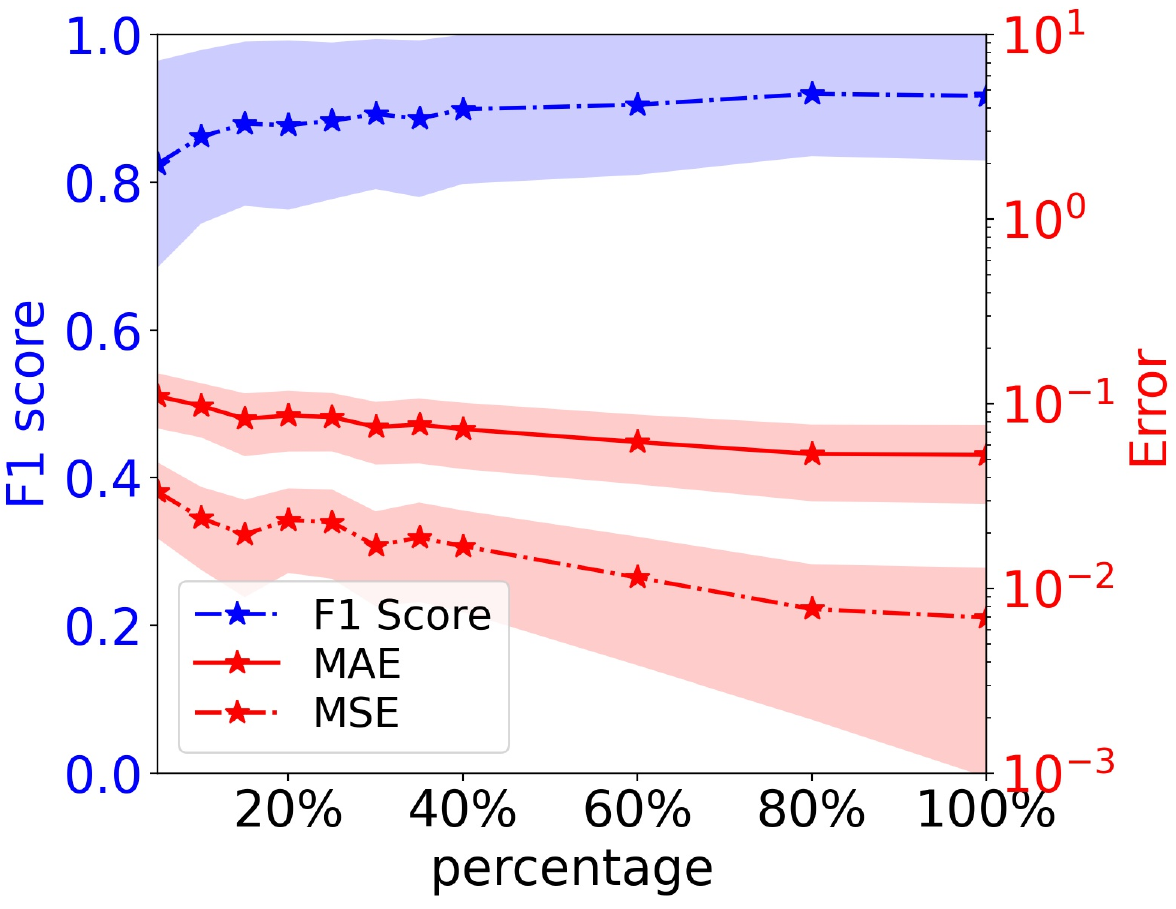}
  \caption{Level 2}
  \label{fig:level2_perturb_metric}
\end{subfigure}
\begin{subfigure}{0.33\linewidth}
  \centering
  \includegraphics[width=\linewidth]{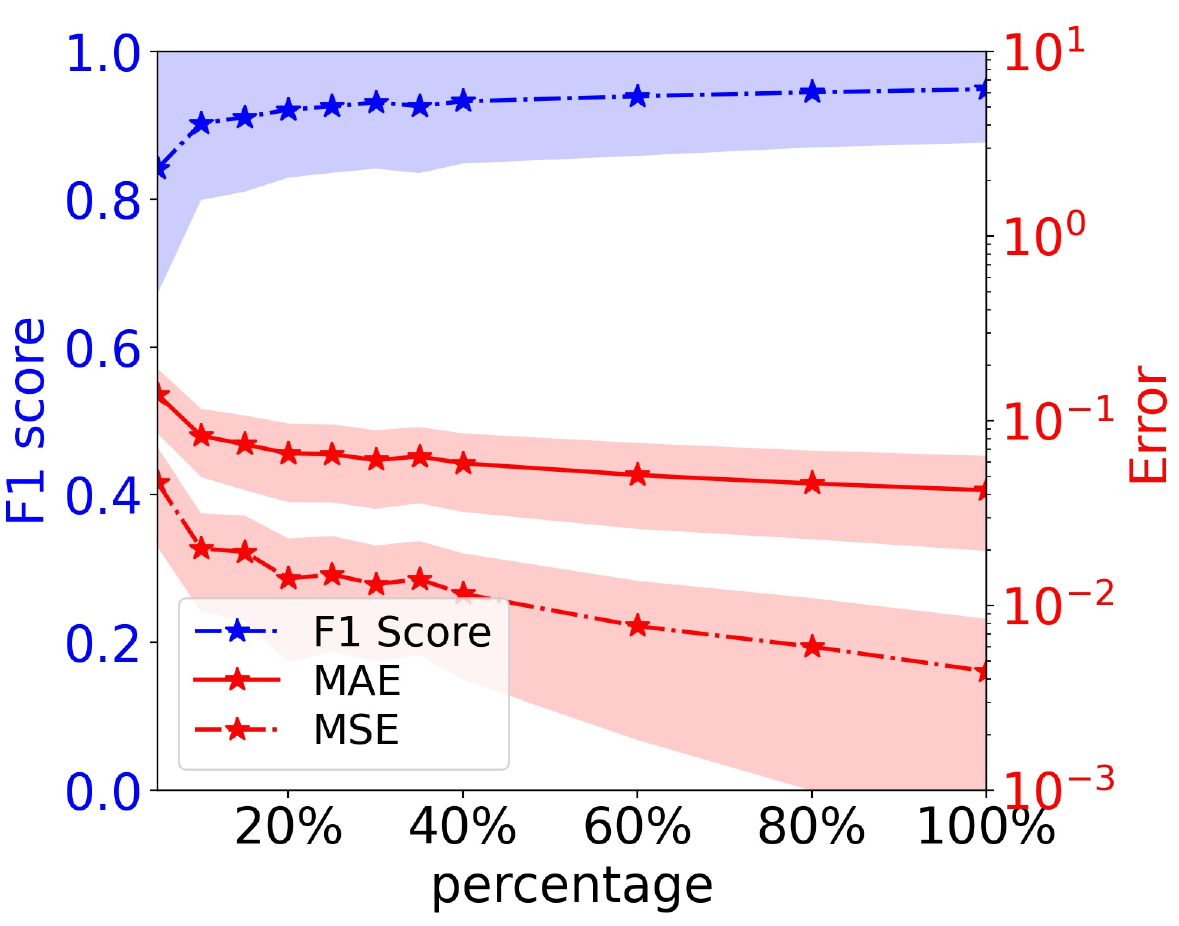}
  \caption{Level 3}
  \label{fig:level3_perturb_metric}
\end{subfigure}
\caption{Relationship between different performance metrics and different ratios of training target nodes, shown for the Level 1 region (Figure \ref{fig:level1_perturb_metric}), Level 2 region (Figure \ref{fig:level2_perturb_metric}), and Level 3 region (Figure \ref{fig:level3_perturb_metric}) with perturbed edge features.}
 % . Right $y$-axis is in logarithmic scale.
\label{fig:metric_perturb}
\end{figure}

\begin{figure}[hbt]
\centering
\begin{subfigure}{0.33\linewidth}
  \includegraphics[width=\linewidth]{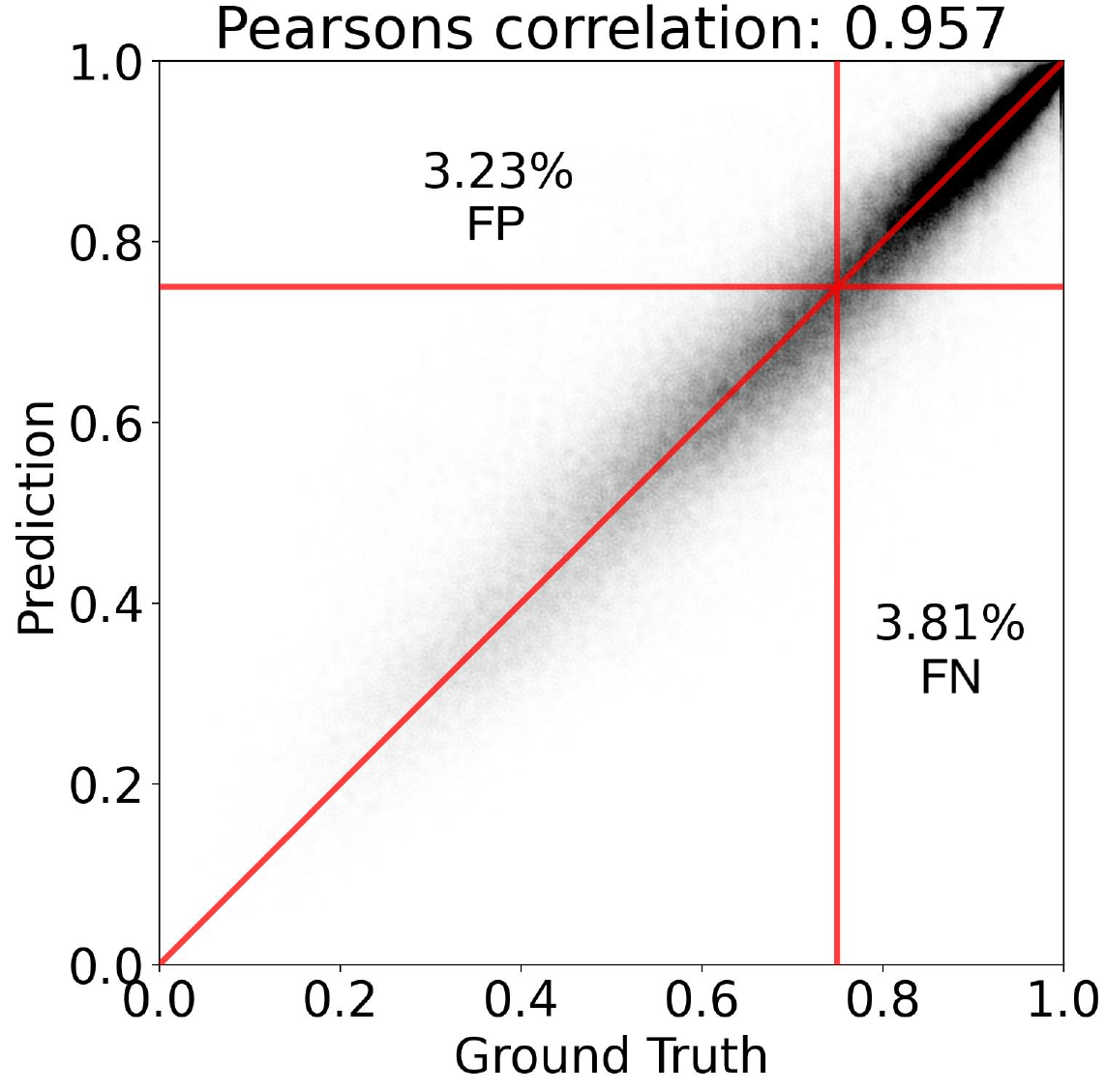}
  \caption{Level 1}
  \label{fig:level1_perturb_corr}
\end{subfigure}
\begin{subfigure}{0.33\linewidth}
  \centering
  \includegraphics[width=\linewidth]{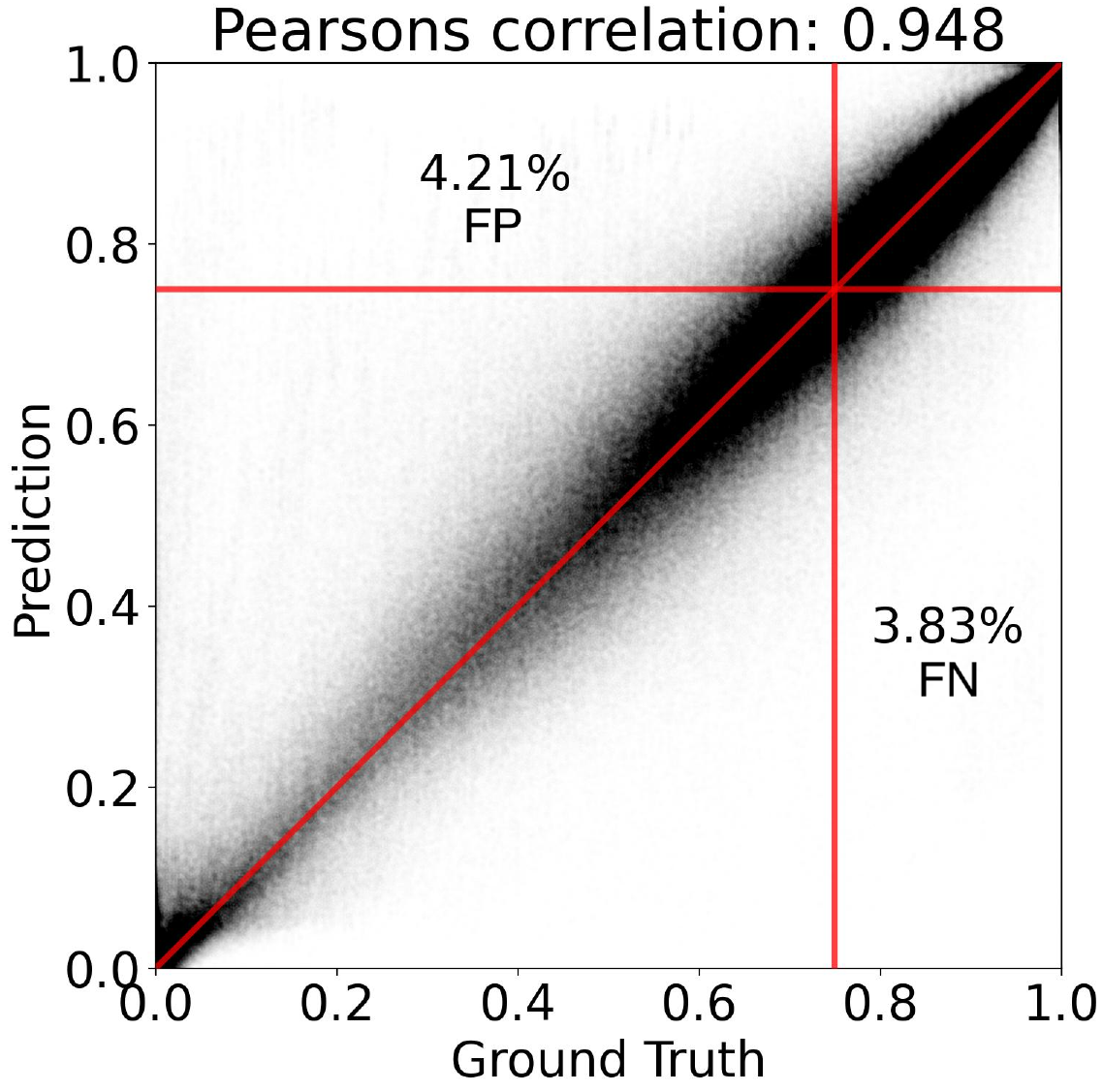}
  \caption{Level 2}
  \label{fig:level2_perturb_corr}
\end{subfigure}
\begin{subfigure}{0.33\linewidth}
  \centering
  \includegraphics[width=\linewidth]{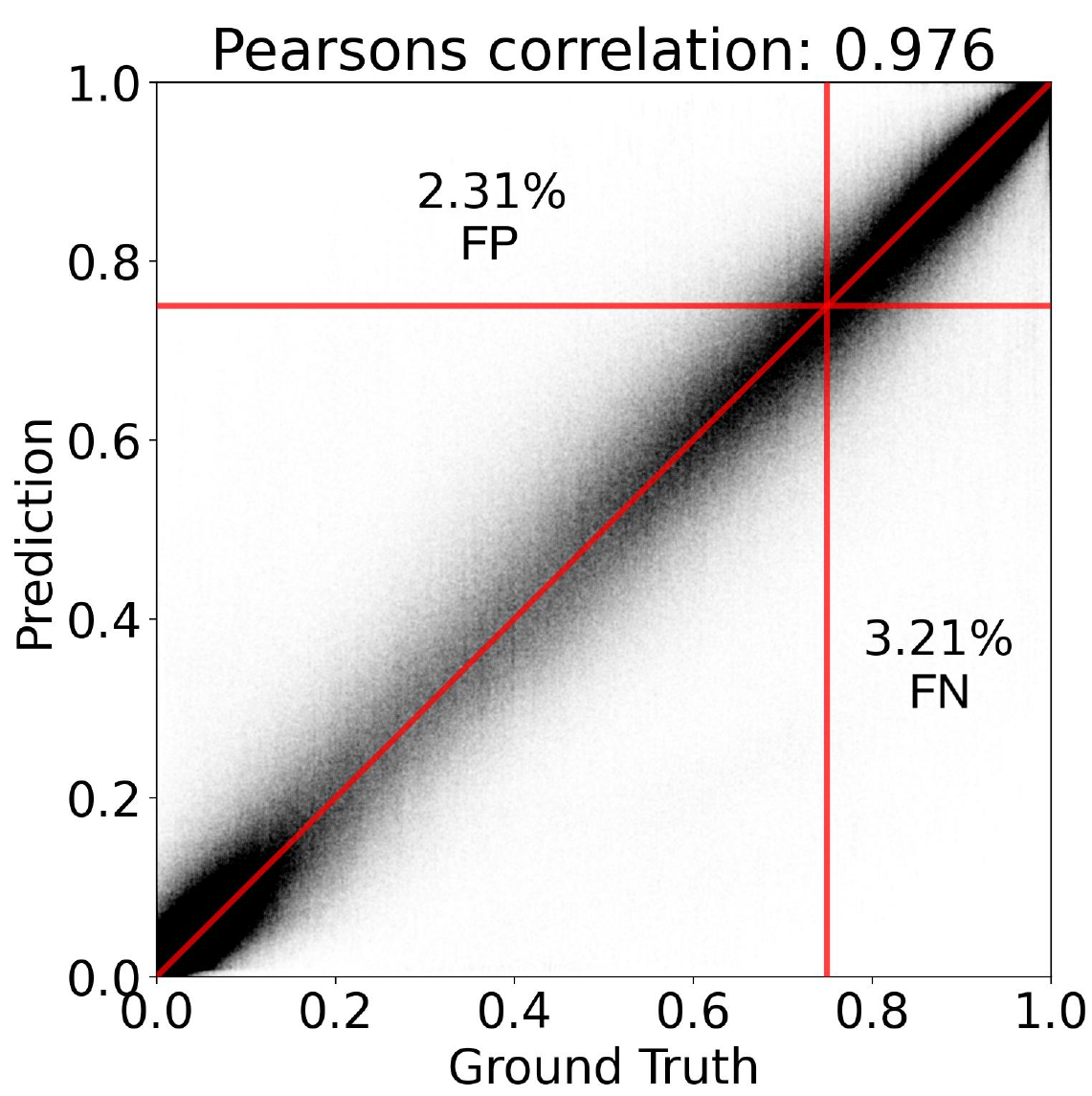}
  \caption{Level 3}
  \label{fig:level3_perturb_corr}
\end{subfigure}
\caption{Comparison of predicted node connectivity between MCS and GNN in Level 1 region (Figure \ref{fig:level1_perturb_corr}), Level 2 region (Figure \ref{fig:level2_perturb_corr}), and Level 3 region (Figure \ref{fig:level3_perturb_corr}) with perturbed edge features.}
% . Diagonal red line represents the line $y=x$. 
\label{fig:corr_perturb}
\end{figure}

% \subsubsection{Model Robustness Performance}
% \label{sec:robustness}
Previous research \cite{xu2020adversarial} indicates that neural networks are vulnerable to small perturbations to the original data, which will might change the neural network prediction drastically. To illustrate the robustness of our approach for reliability analysis, the proposed GNN model is evaluated on the slightly modified testing data. As the context, we consider the cases where bridge failure probabilities are decreased or increased because of repairs or deterioration through the life cycle. To test the robustness of the proposed model among all region levels, we generate perturbed features by adding to each edge failure probability a zero-mean Gaussian noise with a standard deviation set equal to  20\% of the original edge failure probability. The resulting failure probability is truncated at 0 and 1. Following this perturbation to the edge features, the node features are also updated accordingly. Then the GNN prediction of node connectivities is performed for all three regions as described in Section \ref{sec:predictionresult} and the comparison between the prediction and MCS in this perturbed case is also shown in Figures \ref{fig:metric_perturb} and \ref{fig:corr_perturb}. It can be seen in Figure \ref{fig:corr_perturb}, that even though there is more discrepancy between the GNN and  MCS compared to the original case in Figure \ref{fig:corr}, the overall the performance is acceptable for an unseen case where all edges have different features. The F1 score of prediction with perturbed data is 0.964, 0.959, and 0.972 for three region levels, respectively.

Finally, we assess the capability of the proposed approach in generalization or \emph{inductive} inference and prediction. Specifically, we evaluate whether the model can be used for predicting out-of-distribution (OOD) data, that is, producing accurate predictions for test cases that are distributed differently from the way training data was distributed. As opposed to transfer learning, where the model is retrained on the new data, in generalization or inductive learning, we do not modify the previously trained model and only evaluate it on the new data. In practice, this can be used to assess network connectivity when new bridges or roads are either added to a network expansion or removed due to a disaster. 

\begin{figure}[hbt]
\centering
\begin{subfigure}{0.4\linewidth}
  \includegraphics[width=\linewidth]{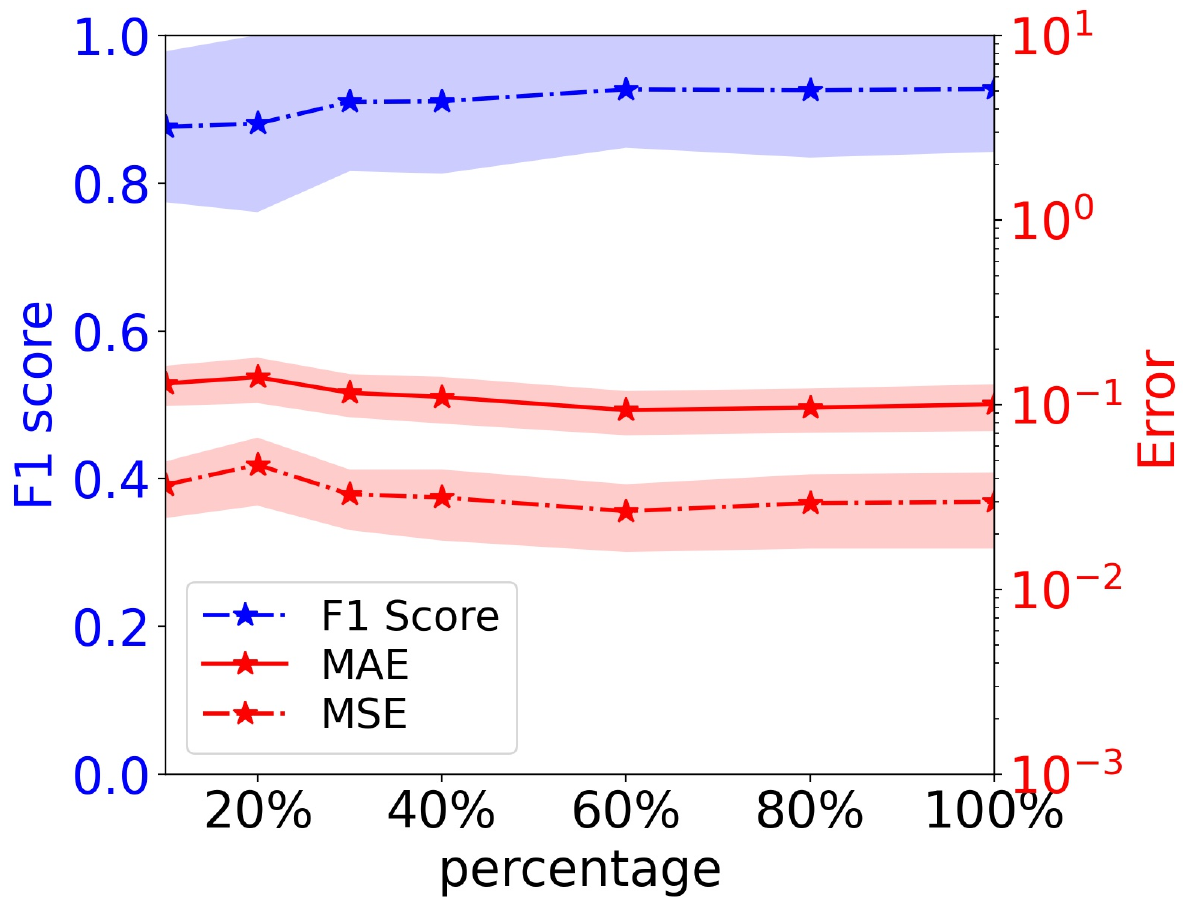}
  \caption{}
  \label{fig:level3_metric_transfer}
\end{subfigure}
\begin{subfigure}{0.32\linewidth}
  \centering
  \includegraphics[width=\linewidth]{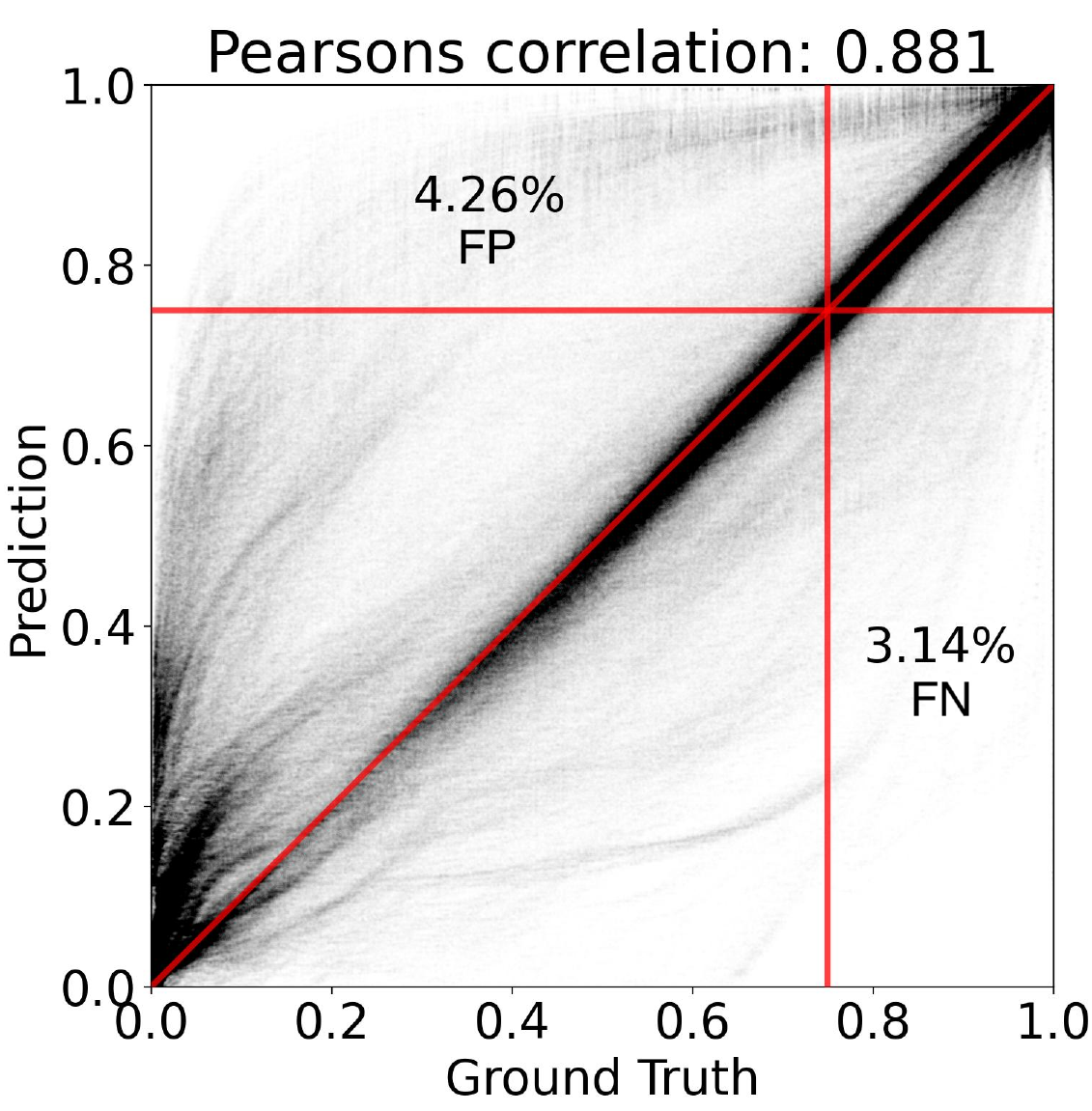}
  \caption{}
  \label{fig:level3_corr_transfer}
\end{subfigure}
\caption{Assessment of the performance of inductive learning from the Level 2 region to Level 3 region, in terms of the classification results (Figure \ref{fig:level3_metric_transfer}) and regression results (Figure \ref{fig:level3_corr_transfer}).}
\label{fig:level3_transfer}
\end{figure}

In these cases, to assess whether the trained model has generalization capacity without additional fine-tuning, we consider the base training of the model to be done at the Level 2 region level. We then test the performance of the trained model on the Level 3 region. In general, this is a challenging task because many of the nodes in the larger Level 3 region are unseen and were not included in the previous training. Furthermore, another challenge in this task is that the input and output dimensions of the neural network change, where the conventional machine learning model cannot handle this scenario. However, as can be seen in Figure \ref{fig:level3_transfer}, the proposed GNN model has an acceptable performance. It can be seen that the F1 score in this case is 0.96 when 80\% of the nodes in the smaller region (Level 2) were used as training nodes, for a prediction at a large region (Level 3). This good performance in handling inductive learning tasks is achieved because the mechanism of the message passing used in the GNN architecture enables a modular understanding of how the graph features are aggregated and collectively influence the task at hand.

\section{Conclusion and Discussion}
\label{sec:conclusion}
Rapid evaluation of large-scale infrastructure system reliability is critical to enhance the preparation, risk mitigation and response management under probabilistic natural hazard events. In this paper, we propose a rapid seismic reliability assessment approach for roadway transportation systems with seismic damage on bridges using the graph neural network. Compared with the common reliability assessment approaches, which consider the response at the system or graph level, we focus on a more detailed notion of reliability where the seismic impact is quantified at the node-level for transportation systems. The proposed GNN surrogate bypasses extensive sampling that is required in Monte Carlo-based approaches and offers high efficiency while preserving accuracy. Additionally, the message-passing component of the proposed GNN model creates a modular model structure that can offer a good generalization capacity, i.e., can effectively predict system reliability in unseen graphs. The numerical experiments demonstrate the effectiveness, robustness, and efficiency of the proposed approach in enabling rapid assessment of large-scale network reliability with high accuracy. As extensions to this work, measures other than node-to-node connectivity can be considered. Examples include travel distance, travel time, or traffic flow. Currently, only the damage beyond the extensive damage is considered in the paper. Another direction for future research is to investigate the influence of multiple damage states in the resulting network level performances, which can provide more insights for the practical implications of the reliability analysis of transportation systems.

\subsection*{Acknowledgments}
This material is based in part upon work supported by the National Science Foundation under Grant No. CMMI-1752302.

\bibliographystyle{unsrt}  
\bibliography{references}  %%% Remove comment to use the external .bib file (using bibtex).

\end{document}